%% file: main.tex
\documentclass{article}

\PassOptionsToPackage{numbers, compress}{natbib}
\usepackage[preprint]{nips2024/neurips_2024}

\usepackage[utf8]{inputenc} % allow utf-8 input
\usepackage[T1]{fontenc}    % use 8-bit T1 fonts
\usepackage{hyperref}       % hyperlinks
\usepackage{url}            % simple URL typesetting
\usepackage{booktabs}       % professional-quality tables
\usepackage{amsfonts}       % blackboard math symbols
\usepackage{nicefrac}       % compact symbols for 1/2, etc.
\usepackage{microtype}      % microtypography
\usepackage{xcolor}         % colors

\usepackage{amsmath}
\usepackage{amsthm}
\newtheorem{theorem}{Theorem}
\newtheorem{assumption}[theorem]{Assumption}

\usepackage{graphicx}
\usepackage{caption}
\usepackage{subcaption}  % for subfigure
\usepackage{algorithm,algorithmic}

\usepackage[algo2e]{algorithm2e} 
\usepackage{listings}
\usepackage{makecell}
\usepackage[capitalise]{cleveref}  % for \cref

\newcommand{\block}{ $1024$ }
\newcommand{\flen}{ $64$ }
\newcommand{\mseg}{ $3$ }

\usepackage{color}

\title{TransformerFAM: Feedback attention is working memory}

\author{%
Dongseong Hwang \quad Weiran Wang \quad Zhuoyuan Huo \quad Khe Chai Sim \quad Pedro Mengibar \\
Google LLC \\
Mountain View, CA, USA \\
\texttt{dongseong@google.com} \\
}

\begin{document}

\maketitle

\input{_1-intro}
\input{_3-method}
\input{_4-experiments}

\input{_2-related}
\input{_5-conclusion}

\bibliographystyle{ksfh_nat}
\bibliography{main}

%%%%%%%%%%%%%%%%%%%%%%%%%%%%%%%%%%%%%%%%%%%%%%%%%%%%%%%%%%%%

\newpage
\appendix
\input{_9-appendix}

%%%%%%%%%%%%%%%%%%%%%%%%%%%%%%%%%%%%%%%%%%%%%%%%%%%%%%%%%%%%
\newpage

\end{document}

%% file: _1-intro.tex
\begin{abstract}
While Transformers have revolutionized deep learning, their quadratic attention complexity hinders their ability to process infinitely long inputs. We propose Feedback Attention Memory (FAM), a novel Transformer architecture that leverages a feedback loop to enable the network to attend to its own latent representations. This design fosters the emergence of working memory within the Transformer, allowing it to process indefinitely long sequences. TransformerFAM requires no additional weights, enabling seamless integration with pre-trained models. Our experiments show that TransformerFAM significantly improves Transformer performance on long-context tasks across various model sizes (1B, 8B, and 24B). These results showcase the potential to empower Large Language Models (LLMs) to process sequences of unlimited length.
\end{abstract}

\section{Introduction}
\label{sec:intro}

The introduction of the Transformer architecture~\cite{vaswani2017attention} has revolutionized deep learning by permeating diverse domains and enhancing performance due to its efficacy and scalability. This scalability fuels a trend analogous to Moore's law, which links increased model size to performance gains~\cite{kaplan2020scaling}.

% \weiran{This paragraph could be significantly shortened or removed if needed.} -> story is about attention success story. 1. attention to homogeneous data and scaling. 2. attention to heterogeneous data. and 3. attention to feedback. this is important part of whote story.
The effectiveness of attention in text sequence processing was solidified through the Transformer paper. Models like BERT~\cite{devlin2018bert} and GPT-3~\cite{brown2020language} further showcased the scalability of Transformer and its tendency for improved performance with increased model size. Following the replacement of LSTM~\cite{sepp1997lstm} by Transformer in most Natural Language Processing (NLP) domains, the Vision Transformer (ViT)~\cite{dosovitskiy2020image} replaced Convolutional Neural Network (CNN)~\cite{lecun1995convolutional} with Transformers in the vision domain, and Conformer (Convolution-augmented Transformer)~\cite{gulati2020conformer} replaced LSTM in the speech domain. The Transformer has become the de facto architecture in various domains. Currently, attention serves as the leading architecture for extracting meaningful representations from homogeneous data.

The logical progression points toward extending attention capabilities to heterogeneous data. This has enabled advances in multimodal fusion (text and vision), as seen in models like DALL·E 2~\cite{ramesh2022hierarchical}, Flamingo~\cite{alayrac2022flamingo} and CoCa~\cite{yu2022coca}. AudioLM~\cite{borsos2023audiolm} has shown that attention also excels at fusing audio and text. Consequently, Gemini~\cite{team2023gemini} integrates text, images, audio, and video into a single generative model. This was possible because attention to heterogeneous data works exceptionally well.

% master slide: https://docs.google.com/presentation/d/1C_WY8jPWiiZRZMGahFtk-0sh-s3D8PQZIWmfdJxx-js/edit#slide=id.g2b19d5fce10_0_378

% https://docs.google.com/drawings/d/1PQoKKd_F0GUR51q6VIrfQtQ77pOM7xBSvnt1uV2FXnA/edit
% https://docs.google.com/drawings/d/1ZEjVPKXjglv8cM_JrD-p0Zhz0oQQvdT_dMOqAI9Zygo/edit
% https://docs.google.com/drawings/d/1RHpDCQdrzaWrSysgmYGzV4UHVpSertLeG6Fu6Zituc4/edit
% https://docs.google.com/drawings/d/18TuFhy7k7LBuU71y9inMPpRf1Cve41_mDyx_pOxNoxQ/edit
\begin{figure*}[t]
    \centering
    \begin{subfigure}{.22\textwidth}
    \includegraphics[width=\linewidth]{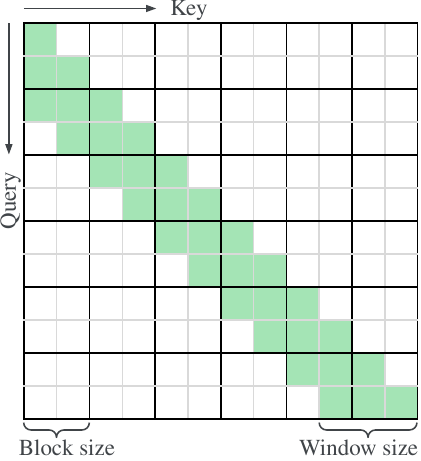}
    \caption{SWA \label{fig:swa}}
    \end{subfigure}\hfill
    \begin{subfigure}{.22\textwidth}
    \includegraphics[width=\linewidth]{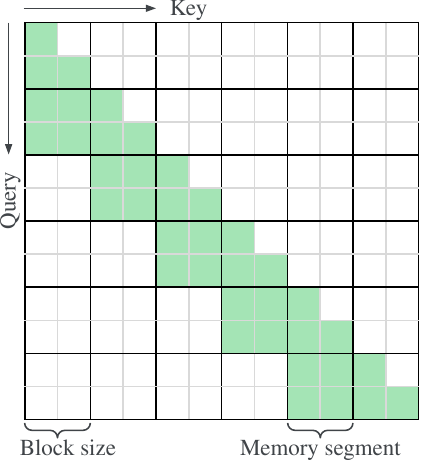}
    \caption{BSWA \label{fig:bswa}}
    \end{subfigure}\hfill
    \begin{subfigure}{.22\textwidth}
    \includegraphics[width=\linewidth]{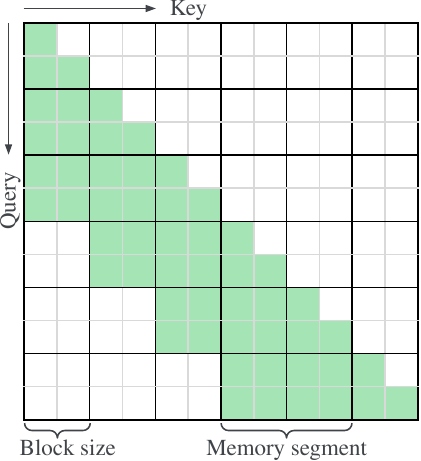}
    \caption{BSWA (2 segments) \label{fig:bswa2}}
    \end{subfigure}\hfill
    \begin{subfigure}{.30\textwidth}
    \includegraphics[width=\linewidth]{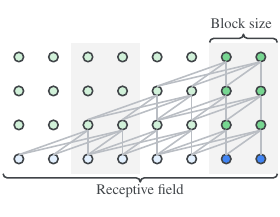}
    \caption{Receptive Field \label{fig:rf}}
    \end{subfigure}
    \hfill
    \caption{ Comparison of query-key attention masks for Sliding Window Attention (SWA) variants. (a) Sliding Window Attention: Attention is restricted to the current window $=3$. (b) Block Sliding Window Attention (BSWA) (block size $=2$, memory segment $=1$): Attention is allowed to previous blocks within the memory segment. (c) BSWA (block size $=2$, memory segment $=2$): The memory segment is expanded, allowing attention to a larger past context. (d) Illustrates the receptive field of BSWA (block size $=2$, memory segment $=1$, depth $=4$): The region within the curly braces represents the receptive field.}
    \label{fig:apndx_my_label}
    % \vspace{-4mm}
\end{figure*}

% https://docs.google.com/drawings/d/1S5KC2fqH7qbI1o_FX9vS-wFfkxywUd4Vw08JEQI3q6M/edit
% https://docs.google.com/drawings/d/1fg5pSuyekRCrhmQI7qx14iUb2nzOCDCEcxwRajIZSlA/edit
\begin{figure*}[thb]
    \centering
    \begin{subfigure}{.35\textwidth}
    \includegraphics[width=\linewidth]{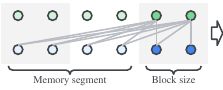}
    \caption{TransformerBSWA \label{fig:transformer_bswa}}
    \end{subfigure}\hfill
    \begin{subfigure}{.65\textwidth}
    \includegraphics[width=\linewidth]{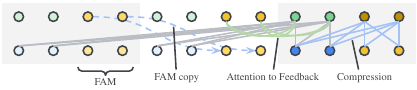}
    \caption{TransformerFAM \label{fig:transformer_fam}}
    \end{subfigure}\hfill
    \caption{ Comparison of attention patterns in Transformer layer. (a) TransformerBSWA: Input query attends to the current block and two memory segments, providing past context. (b) TransformerFAM: Input query attends to the current block, memory segments, and past FAM (green lines). FAM query (copied from previous FAM, blue dash arrow) compresses the current block to update FAM. This feedback loop enables information compression and propagation over indefinite horizon, which is working memory. \cref{fig:animation} shows in detail how the dynamic process occurs over time.}
    % \vspace{-4mm}
\end{figure*}

% \weiran{, which are essential in document summarization / question answering ...} -> it's not only for document summarization.
Despite the impressive success of attention, it suffers from major drawbacks. Firstly, attention has quadratic complexity with respect to context length, which limits the capability of modeling long contexts. Secondly, it forgets information from context before attention window, unlike LSTM, which theoretically can propagate information indefinitely. We want the better architecture to be able to process arbitrarily long sequences efficiently, while preserving very long-term dependencies.
% \weiran{You could add sth like this here to transit to your contributions: ``In this work, we propose a novel architecture that tackle both drawbacks, motivated by the working memory theory from neuralscience.''} -> it will say that later.

% \weiran{I feel it may be better to distribute this paragraph and references to the previous paragraph when you talk about each of the drawbacks. The sentiment is that ``although there have been prior works attempting to address the drawbacks, they still suffer from performance degradation or their approaches are ad-hoc, blah blah".} -> this paragraph provides the solutions other papers suggested because of the problem of previous paragraph.
Sliding window attention is introduced~\cite{dai2019transformer,beltagy2020longformer} to handle infinitely long sequences as input. However, it disregards information beyond the effective receptive field (approximately model depth $\times$ window size in \cref{fig:rf}) during sequence generation. Various approaches have attempted to address the long-context problem by sparse attention~\cite{child2019generating,beltagy2020longformer,zaheer2020big,kitaev2020reformer,roy2021efficient,oren2024transformers} and linear approximated attention~\cite{wang2020linformer,katharopoulos2020transformers,choromanski2020rethinking,xiong2021nystromformer}, showing effectiveness below the 1B scale. Yet, scaling laws~\cite{tay2022scaling} suggest these approximations do not perform as well at GPT-3 level. Current SoTA large language model (LLM) leaderboards~\cite{llmleaderboard} do not feature architectures primarily relying on approximated attention.

On the other hand, Neuroscience links attention to multisensory integration~\cite{tang2016interactions}. Endogenous (goal-driven) and exogenous (stimulus-driven) attention are distributed throughout sensory processing regions in brain, and the brain areas correlated with the attention overlap substantially with the multisensory cortical regions involved in multisensory integration. Working memory exhibits similar distribution~\cite{christophel2017distributed}, overlapping with attention-correlated areas. This indicates that attention is an important factor in not only multimodal fusion but also working memory.

In the human brain, working memory~\cite{wallace1960plans} provides a temporary memory for performing tasks. While working memory is stored in sustained activations, long-term memory is stored in weights~\cite{fuster1973unit}. LLMs has enormous long-term memory enough to store the entire internet~\cite{villalobos2022will}, but they do not have working memory. The activation of working memory is sustained by prefrontal cortical-thalamic loops~\cite{ashby2005frost}, which means working memory is sustained by the continuous spiking of activation within a feedback loop.

% \weiran{Try to strengthen the theory on working memory with your actual algorithm, what are the important properties of working memory thathelp us tackel of the drawbacks of standard transformers, or how is your algorithm design, e.g., feedback loop, inspired by it?} -> Done

% We desire to incorporate working memory into the Transformer.
% Equipped with working memory, a Transformer would alleviate the need to attend to extremely long contexts. It enables the compressed important information from the input sequence despite its length, enhancing tasks requiring long-range dependency understanding.
% \weiran{Introduce the name of your method here? but this paragraph is somewhat redundant as it overlaps with the 2nd last paragraph of Section 1.} -> it's introduced later.

Attention has been shown to be effective for processing both homogeneous and heterogeneous data. The next natural step is to apply attention to its own latent representations through a feedback loop. We hypothesize that this next step will naturally lead to the emergence of working memory in Transformers.

\begin{assumption}
The attention mechanism within the feedback loop functions as working memory.
\label{ass:feedback_memory}
\end{assumption}

Feedback connections are prevalent in biological neural networks. Even organisms with simple neural structures, such as C. elegans (with only 302 neurons)~\cite{white314s}, exhibit various feedback loops, like connections from higher-level interneurons to lower-level ones~\cite{hasani2018can}. However, incorporating feedback loops in Transformer is challenging. There are two main approaches. The first approach is linking the topmost layer to the bottommost~\cite{fan2020addressing,bulatov2022recurrent}. However, this cannot model feedback between interneurons, and this has only one global working memory. The second approach is within-transformer-block feedback: the output activation of a Transformer layer is fed back as input to the same layer. This is the approach we propose, which enables each Transformer layer to have a distributed working memory that corresponds to its abstraction level.
% \weiran{but this feedback happens across two consecutive time steps?} -> I don't fully catch up what's question?

Recurrent Neural Networks (RNNs) have achieved great success in machine learning by introducing feedback loops~\cite{hochreiter1997long,cho2014learning}. RNNs pass feedback between sequences through hidden states. Attention mechanisms can implement feedback loops by attending to both the input sequence and the feedback state simultaneously.

% \weiran{Move the relevant figures here to better illustrate your proposal.} -> good idea. Let me try when paper is finalized. Now it causes bloated layout.
We propose a novel Transformer architecture (TransformerFAM) that enables attention to both homogeneous sequence data and latent representations via a feedback loop. This architecture change fosters the natural emergence of working memory within Transformers. During inference, TransformerFAM has a computational complexity of $O(L)$ and a memory complexity of $O(1)$, where $L$ is the length of the processed tokens. TransformerFAM can maintain past information for an indefinite horizon, making it a promising solution for LLMs to handle infinitely long input sequences.
%  \weiran{don't you need to list the block size as a parameter? also the complexity is still quadratic within the block, right?} -> O(LxB). it's explained in method section.

TransformerFAM does not introduce new weights to the Transformer, allowing the reuse of pre-trained checkpoints. Our experiments show that fine-tuning TransformerFAM with LoRA for just 50k steps significantly enhances performance on long-context tasks across 1B, 8B, and 24B Flan-PaLM LLMs~\cite{chung2022scaling}.  

%% file: _3-method.tex
\section{TransformerFAM}
\label{sec:xformerfam}

\subsection{Block Sliding Window Attention (BSWA)}
\label{ssec:bswa}

In general, when handling long context inputs, there are the two main approaches. A first approach is to increase the context length with increasing the computational resources (memory and processing power). A second approach is the implementation of Sliding Window Attention (SWA)~\cite{beltagy2020longformer,jiang2023mistral}, as illustrated in \cref{fig:swa}. During inference, the standard implementation allocates key and value cache twice the window length at the beginning, and then using a ring buffer to update the necessary components at each step, in order to avoid memory allocation and copying operations every step, which are computationally expensive.

Longformer~\cite{beltagy2020longformer} introduced Sliding Window Attention, which caches on a block-by-block basis. We will refer to this as Block Sliding Window Attention (BSWA). BSWA does not mask out past keys and values in the ring buffer, but attends to all the information in the ring buffer.

BSWA has two hyperparameters: block size and memory segment, as illustrated in Figure~\ref{fig:bswa} and \ref{fig:bswa2}. Block size determines how many tokens are in each block and is also used as the stride for sliding. Memory segment determines how many past blocks to cache. In our experiments, we set the default hyperparameters as follows: block size of \block and memory segment of \mseg (corresponding to a window size ranging from 3073 to 4096).

Given a sequence input $I = [i_1, i_2, \dots, i_T]$ to a vanilla Transformer layer, Transformer transforms the input to a sequence output $O = [o_1, o_2, \dots, o_T]$ as follows~\cite{vaswani2017attention}:

\begin{align*}
Q, K, V &= \text{QKV}(\text{PreLN}(I)) \\
a_t &= \text{SelfAttention}(q_t, K, V) + i_t \\
O &= \text{FF}(\text{PreLN}(A)) + A
\end{align*}

While PreLN~\cite{xiong2020layer} is used in the equation, it is not mandatory. Autoregressive Transformer changes the SelfAttention equation as follows:
$a_t = \text{SelfAttention}(q_t, K_{:t}, V_{:t}) + i_t$.

% TODO(dongseong)
% \begin{align*}
% a_t &= \text{SelfAttention}(q_t, K_{:t}, V_{:t}) + i_t \\
% \end{align*}

In a Transformer with SWA, the equation is modified to account for a window size of $w$: $a_t = \text{SelfAttention}(q_t, K_{t-w:t}, V_{t-w:t}) + i_t$.

% TODO(dongseong)
% \begin{align*}
% a_t &= \text{SelfAttention}(q_t, K_{t-w:t}, V_{t-w:t}) + i_t \\
% \end{align*}

In a Transformer with BSWA, $\tau$ denotes the block index. Each block $\tau$ contains a set of keys and values, determined by the block size. The equation is modified to account for a block index $\tau$ and a memory segment $m$ in \cref{eq:bswa}. $K_{\tau - m:\tau-1}$ is the keys in the memory segments from the $m$ blocks before $\tau$ to the block before $\tau$, and $K_{\tau,:t}$ is the keys from the beginning of $\tau$ block up to $t$. We will refer to Transformer with BSWA as TransformerBSWA.

\begin{subequations}
\begin{align}
\hat{K}_t &= \text{Concat}(K_{\tau - m:\tau-1}, K_{\tau,:t}) \\
\hat{V}_t &= \text{Concat}(V_{\tau - m:\tau-1}, V_{\tau,:t}) \\
a_t &= \text{SelfAttention}(q_t, \hat{K}_t, \hat{V}_t) + i_t 
\end{align}
\label{eq:bswa}
\end{subequations}

\cref{alg:bswa} presents TransformerBSWA (\cref{eq:bswa}), re-expressed from the perspective of a block index $\tau$. \cref{alg:bswa} describes how to iteratively calculate $a_t$ and then concatenate the results into an $A_\tau$ sequence. Typically, standard implementations employ a causal attention mask to enable parallel computation of self-attention.

TransformerXL~\cite{dai2019transformer} proposed to use a technique called "stop gradient" for the memory segment. However, we argue that this technique has a negative impact on ability of the model to attend to past information. Specifically, we show that using stop gradient results in a much shorter receptive field than the theoretical receptive field in \cref{ssec:passkey}. We believe that allowing gradient to flow to the memory segment is necessary for the model to learn to carry important information in the memory segment.

% https://tex.stackexchange.com/questions/403823/how-to-use-function-in-latex-algorithm
\begin{algorithm}[H]
\SetAlgoLined
\DontPrintSemicolon
\KwIn{$ I_{\tau}, K_{\tau - m:\tau-1}, V_{\tau - m:\tau-1} $}    
\KwOut{$ O_{\tau} $}
    \SetKwFunction{FMain}{Xformer}
    \SetKwProg{Fn}{Function}{:}{}
    \Fn{\FMain{$I_{\tau}, K_{\tau - m:\tau-1}, V_{\tau - m:\tau-1}$}}{
        $Q_{\tau}, K_{\tau}, V_{\tau} \gets \text{QKV}(\text{PreLN}(I_{\tau}))$
        
        $\hat{K}_t \gets \text{Concat}(K_{\tau - m:\tau-1}, K_{\tau,:t})$
        
        $\hat{V}_t \gets \text{Concat}(V_{\tau - m:\tau-1}, V_{\tau,:t})$
        
        $a_t \gets \text{SelfAttention}(q_t, \hat{K}_t, \hat{V}_t) + i_t$
        
        $O_{\tau} \gets \text{FF}(\text{PreLN}(A_{\tau})) + A_{\tau}$

        \textbf{return} $ O_{\tau} $ 
}
\textbf{End Function}
\caption{The function of TransformerBSWA}
\label{alg:bswa}
\end{algorithm}

\begin{algorithm}[H]
\SetAlgoLined
\DontPrintSemicolon
\KwIn{$ I_{\tau}, K_{\tau - m:\tau-1}, V_{\tau - m:\tau-1}, F_{\tau-1} $}    
\KwOut{$ O_{\tau}, F_{\tau} $}
    \SetKwFunction{FMain}{Xformer}
    \SetKwProg{Fn}{Function}{:}{}
    \Fn{\FMain{$I_{\tau}, K_{\tau - m:\tau-1}, V_{\tau - m:\tau-1}, F_{\tau-1}$}}{
        $Q_{\tau}, K_{\tau}, V_{\tau} \gets \text{QKV}(\text{PreLN}(I_{\tau}))$
        
        $Q^F_{\tau-1}, K^F_{\tau-1}, V^F_{\tau-1} \gets \text{QKV}(\text{PreLN}(F_{\tau-1}))$
        
        $\hat{K}_t \gets \text{Concat}(K_{\tau - m:\tau-1}, K^F_{\tau-1}, K_{\tau,:t})$
        
        $\hat{V}_t \gets \text{Concat}(V_{\tau - m:\tau-1}, V^F_{\tau-1}, V_{\tau,:t})$
        
        $a_t \gets \text{SelfAttention}(q_t, \hat{K}_t, \hat{V}_t) + i_t$
        
        $O_{\tau} \gets \text{FF}(\text{PreLN}(A_{\tau})) + A_{\tau}$
        
        $\tilde{K}_{\tau} \gets \text{Concat}(K^F_{\tau-1}, K_{\tau,:t})$
        
        $\tilde{V}_{\tau} \gets \text{Concat}(V^F_{\tau-1}, V_{\tau,:t})$
        
        $A^F_{\tau} \gets \text{SelfAttention}(Q^F_{\tau-1}, \tilde{K}_{\tau}, \tilde{V}_{\tau}) + F_{\tau-1}$
        
        $F_{\tau} \gets \text{FF}(\text{PreLN}(A^F_{\tau})) + A^F_{\tau}$

        \textbf{return} $ O_{\tau}, F_{\tau} $ 
}
\textbf{End Function}
\caption{The function of TransformerFAM}
\label{alg:fam}
\end{algorithm}

% \weiran{This algorithm could be removed if needed.} -> it's helpful to understand algo2. Let me decide it later depending on 8 page limit.

While this modification might seem to burden training memory and computation, it does not significantly impact performance in practice. This is primarily due to the prevalence of gradient checkpointing~\cite{chen2016training} in LLM training on ML accelerators, as memory often presents the primary bottleneck. Gradient checkpointing recomputes attention  during backpropagation, from later blocks to earlier blocks. Therefore, the presence or absence of stop gradients has little impact on the overall computational complexity, while still improving performance.

In addition, when training an LLM with sliding window attention on long context inputs (more than $8$k tokens), computing the attention over the entire input at once would require too much memory. As a result, the standard practice is to divide the attention into blocks and calculate it using a vectorized map (e.g., jax.vmap, torch.vmap). This reduces the peak memory usage to the amount required to calculate one block. Blocks are independent of each other, so they can be calculated in any order. Understanding this point is essential when evaluating the memory requirements and training efficiency of Feedback Attention Memory (FAM), as discussed in \cref{ssec:fam}.

Transformer has quadratic memory and computation complexity with respect to the length of the input sequence due to self-attention. It has $O(L^2)$ complexity for input length $L$. However, Transformer with Sliding Window Attention has linear complexity with respect to the input sequence. It has $O(L \times W)$ complexity for input length $L$ and window size $W$. If the input length is about the same as the window size ($\sim 1$k), the complexity difference is almost negligible, but if the input length is large like $128$k in GPT-4 turbo~\cite{achiam2023gpt}, there is a huge difference. In inference, Transformer with SWA or BSWA only needs to cache a fixed ring buffer (block size $+$ memory segment). Therefore, it only consumes constant memory regardless of the generated token length. Therefore, LLMs using SWA or BSWA can generate infinitely long output tokens.

However, BSWA has a limited receptive field, approximately equal to model depth $\times$ window size as illustrated in \cref{fig:rf}. As a result, the later generated tokens are not related to tokens outside the receptive field (e.g., prompt). To address this limitation, we propose a novel architecture in the following \cref{ssec:fam}. Our approach, Feedback Attention Memory (FAM), builds upon BSWA. This is because the block stride concept of BSWA is well-suited for blockwise feedback updates.

% \weiran{I feel this paper could use similar level of details as the Flash Attention paper, as both are careful/efficient implementations.} -> BSWA is not main contribution of this paper, and we already spare too much space for it.

\subsection{Feedback Attention Memory}
\label{ssec:fam}

% \weiran{I would use equations like those in (1a), (1b), (1c) to help understanding, you could use color to emphasize the additional memory component. Algorithm 2 can be moved to appendix.} -> move algo2 near algo1 for easy comparison.

As mentioned in \cref{sec:intro}, we hypothesized that attending to the feedback loop can give rise to working memory in \cref{ass:feedback_memory}. To implement the feedback loop, we add feedback activations that feed contextual representation back into each block of BSWA. We call these virtual activations as Feedback Attention Memory (FAM). FAM is designed to meet the following key requirements:

\begin{itemize}
% \item Self-attention of the Transformer should attend to both the context and FAM simultaneously.
% \item FAM should carry global contextual information over indefinite horizon.
% \item FAM should be updated when transitioning from block $\tau$ to block $\tau+1$.
% \item FAM update compresses the current block $\tau$ information, conditioned on the past FAM $\tau-1$.
\item Integrated Attention: Self-attention should simultaneously process input context and FAM.
\item Block-Wise Updates: FAM should be updated when transitioning between blocks.
\item Information Compression: FAM updates should compress current block information, conditioned on previous FAM.
\item Global Contextual Storage: FAM should store comprehensive contextual information indefinitely.
\end{itemize}

The proposed architecture achieves this by appending FAM to block segments and incorporating it into self-attention processes. This enables richer representations and dynamic propagation of global contextual information across blocks, as illustrated in \cref{fig:transformer_fam}. When self-attention occurs on the current block, the input query for the block attends to the input key for that block, the memory segment, and the previous FAM. The previous FAM provides global contextual information, allowing for a much richer representation than BSWA. In parallel, the FAM query attends to the current block and the FAM key. The FAM query compresses the current block, conditioned on the previous global contextual information. The FAM query is dynamically generated based on previous global contextual information, as it is copied from the previous FAM. Then, the newly updated FAM serves to propagate global contextual information to the next block recursively. This process is formally described in \cref{alg:fam}.

% \weiran{remove or rephrase: but this is because it is written out in a way that makes it easier to understand.} 
While \cref{alg:fam} might initially suggest a doubling of matrix operations compared to \cref{alg:bswa}, it performs the same number of matrix operations in the actual implementation, because it starts with the concatenation of block input $I_{\tau}$ and FAM $F_{\tau-1}$. The attention mask within self-attention requires a minor modification to accurately represent FAM. The FAM $F_{\tau-1}$ is much shorter than the input $I_{\tau}$, and in \cref{sec:exp}, we experimented with a block size of \block and a FAM length of \flen.

Transformers are much better at exploiting the parallelism of ML accelerators than Recurrent Neural Networks (RNNs). This is because RNNs have a causal relationship between input sequences, while Transformers only have a causal relationship between the inputs and the layer one depth below. It is possible to worry that the feedback mechanism of TransformerFAM will eliminate the advantages of Transformers and make training inefficient. As explained in the implementation of BSWA, memory-efficient implementations perform self-attention in blocks using vectorized maps. Otherwise, peak memory increases during LLM training, requiring more ML accelerators. The causal relationship of TransformerFAM only exists between blocks. Since vectorized maps are used to perform self-attention in blocks, the causal relationship between blocks does not affect training speed and memory. In addition, processing \flen additional FAM when processing \block block input sequences has only a minor impact on performance. Therefore, the memory consumption and training speed of TransformerFAM are almost the same as those of TransformerBSWA.

% \weiran{Remove or rephrase this paragraph.}
% FIXME: Remove it and move Appendix here for arxiv paper.
TransformerFAM requires additional considerations for FAM initialization and length extrapolation. These details are explained in \cref{app:additional_fam}.

An evaluation of multiple FAM variants was conducted, and the best-performing variant is presented in the main paper. \cref{app:dont} provides further details for the remaining variants.

%% file: _4-experiments.tex
\section{Experiments}
\label{sec:exp}

\subsection{Training}
\label{ssec:training}

Pretraining an LLM from scratch requires a huge amount of resources. TransformerFAM can reuse existing LLM checkpoints because it does not add new weights to the Transformer layer. We reused 1B, 8B, and 24B Flan-PaLM LLMs~\cite{chung2022scaling} for our experiments. This is a large enough size to prove that TransformerFAM is a general solution for LLMs. The model sizes 1B, 8B, and 24B refer to the size of the plain Transformer, excluding the text embedding table. The models use a 256k sentence piece tokenizer~\cite{kudo2018sentencepiece}, resulting in 400M text embedding table weights for the 1B model and 1B weights for the 8B and 24B models. The detailed model architecture is described in \cref{table:llm_arch} in \cref{app:flan}.

Flan-PaLM is a model that is fine-tuned on top of a pretrained PaLM model~\cite{chowdhery2023palm} using instruction finetuning. The instruction data consists of few-shot instructions with $100$ to $1$k tokens, which are packed into $2.5$k tokens for training. This means that individual instruction data are concatenated until they reach $2.5$k tokens.

We applied both the TransformerBSWA and TransformerFAM architectures to Flan-PaLM and fine-tuned it for an additional 50k steps. We experimented with different memory segment for both architectures. The block size is set to \block and the FAM length is set to \flen.

During fine-tuning, we used the same Flan instruction data packed into $8.5$k tokens. To maintain a minibatch size of 128 for all models, we used 32 TPUv5~\cite{jouppi2023tpu} cores for the 1B model, 64 cores for the 8B model, and 128 cores for the 24B model. If we had used more resources and a larger minibatch size, we might have achieved better results than those reported in the paper.

We performed LoRA finetuning by adding LoRA~\cite{hu2021lora} to the Attention and FF layers of Transformer without training all the parameters. Full finetuning resulted in lower scores on various tasks reported by GPT-3~\cite{brown2020language}, because catastrophic forgetting~\cite{kirkpatrick2017overcoming} occurred in domains that were not covered by the instruction data. In LoRA finetuning, the scores on GPT-3 tasks actually improved, and the performance on long context tasks was similar to that of full finetuning. The rank of LoRA was 64, and the weights of the original Attention and FF layers were merged with LoRA weights and used during inference.

The Adafactor optimizer ( $ \beta_1 = 0.9, \beta_2 = 0.99 $ ) ~\cite{shazeer2018adafactor} was used with constant learning rate. The learning rates used were $10^{-4}$ for 1B, and $3 \times 10^{-3}$ for both 8B and 24B.

In addition, TransformerXL exhibits comparable performance to TransformerBSWA. The implementations are almost identical, with TransformerXL employing an additional QK attention mask to mask out keys beyond a predetermined window size. \textbf{\cref{sssec:txl} demonstrates that the performance difference between TransformerXL and TransformerBSWA is insignificant, and therefore, experimental results for TransformerBSWA are only included in the main paper}.

\subsubsection{Data}
\label{sssec:data}

The ideal data for training TransformerFAM is a very long document with continuous context, such as textbooks and novels, and the data should be large enough to finetune an LLM. Additionally, the same very long document should be used continuously in each minibatch component, while maintaining FAM and memory segments between training steps.

% \weiran{What is the significance of the paragraph? Are you performing next word prediction?} -> it explain current LLM data is not suitable to train memory because of iid.
The loss function of an LLM is to minimize the difference between the parametric probabilistic model $P(x | \theta)$ and the data-generating true distribution $p_{data}(x)$. The Kullback–Leibler divergence (KL divergence) is used to measure this difference. To perform an unbiased estimation of KL divergence, we draw samples from the data-generating true distribution, which are assumed to be independent and identically distributed (IID). However, the ideal training scenario for the aforementioned memory training directly contradicts the IID assumption. We refer to this as the curse of IID.

Due to the curse of IID, we could not find the training infrastructure or data suitable for training memory. So we used Flan instruction data as a last resort.

% \weiran{It may be better to either explain this detail more carefully, because not all readers understand the effect of ``packed segment", or you could omit this detail.} -> already explained above as mentioned: The instruction data consists of few-shot instructions with $100$ to $1$k tokens, which are packed into $2.5$k tokens for training. This means that individual instruction data are concatenated until they reach $2.5$k tokens.
We used Flan instruction data~\cite{weifinetuned_data} as training data, packed up to 8.5k tokens. In the Flan paper~\cite{wei2021finetuned,chung2022scaling}, a special attention mask was used to prevent attention between different segments during self-attention, by applying a separate mask to each packed segment. We did not use this special attention mask processing. Attention occurs causally within the window, regardless of the segment. We expected TransformerBSWA and TransformerFAM to learn to remember important information and forget outdated information by themselves.

Each token in Flan data has a weight. Few shots examples and instructions have a weight of 0, and answers have a weight of 1. This means that the model only learns to generate the answer. To incentivize the model to remember long contexts, we randomly selected 256 consecutive tokens from 8.5k tokens and appended them to the end of the data with the prompt '\texttt{[repeat random segment]:}'. The repeated tokens were given a weight of $0.1$. We hope that future studies can use more suitable data for training memory, such as long continuous documents, long-form speech, video or video game.
% We do not think that this training data is ideal to train the memory capability. We plan to use better data in future studies.

\subsection{PassKey Retrieval}
\label{ssec:passkey}

The PassKey retrieval task is a recent benchmark used in several long-context transformer papers~\cite{mohtashami2023landmark,tworkowski2023focused,chen2023extending}. In this task, a passkey is presented at the beginning, followed by a very long filler context. Finally, a question about the passkey is asked, as shown in \cref{fig:passkey-format} in \cref{app:passkey}.

This task is a good smoke test to quickly check if information is transmitted in a long context. However, this task only checks if small and important information is transmitted, and does not check if large amounts of information can be efficiently compressed.

% \weiran{Consider presenting ablation study on block size if you have them.} -> we don't have ablation study of block size, as avaliable TPUs for speech team is very limited.
We fine-tuned the Flan-PaLM 1B model for 5k steps with the PassKey format, which has a filler context of 2k to 18k randomly. We used a block size of 1024, TransformerBSWA with 0 to 12 memory segments, TransformerFam with 0 memory segments, and a FAM length of \flen. When the number of memory segments is 3, the window size is 3k (i.e. memory segment $ \times $ block size).

\begin{figure*}[thb]
    \centering
    \begin{subfigure}{.5\textwidth}
    \includegraphics[width=\linewidth]{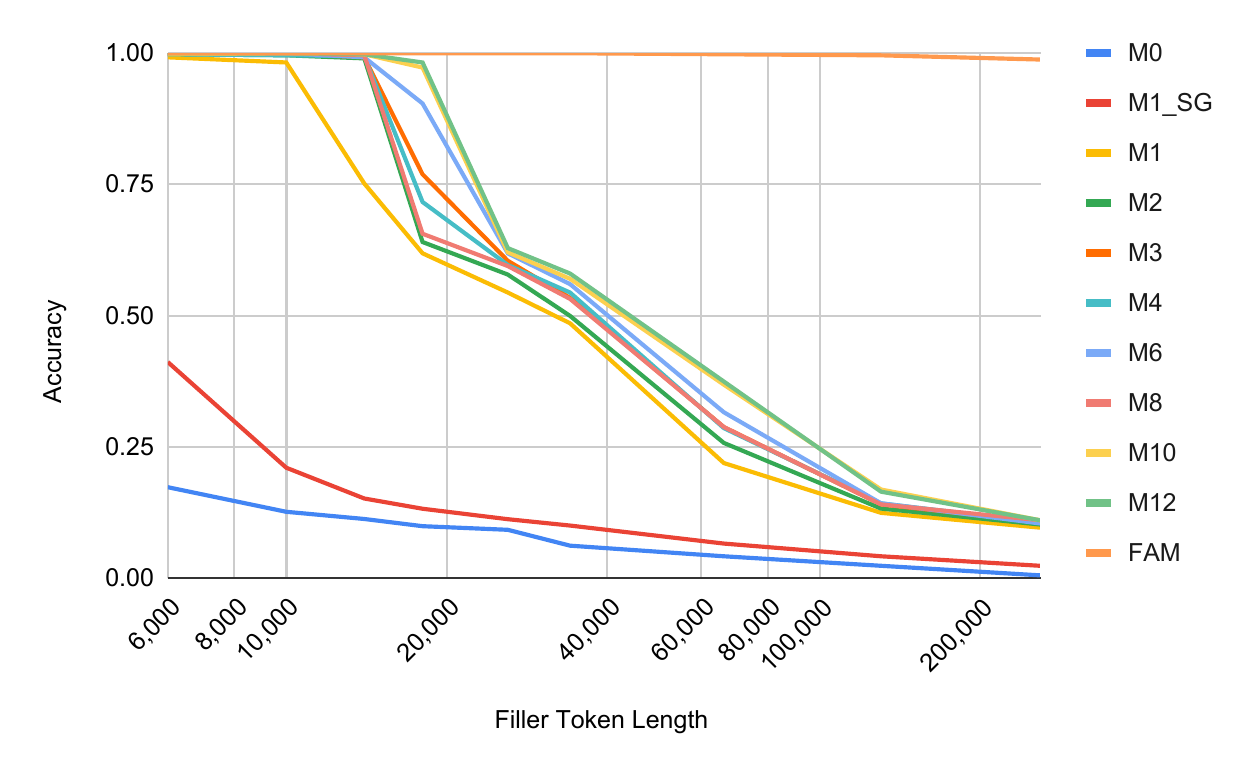}
    \caption{PassKey Retrieval \label{fig:passkey}}
    \end{subfigure}\hfill
    \begin{subfigure}{.5\textwidth}
    \includegraphics[width=\linewidth]{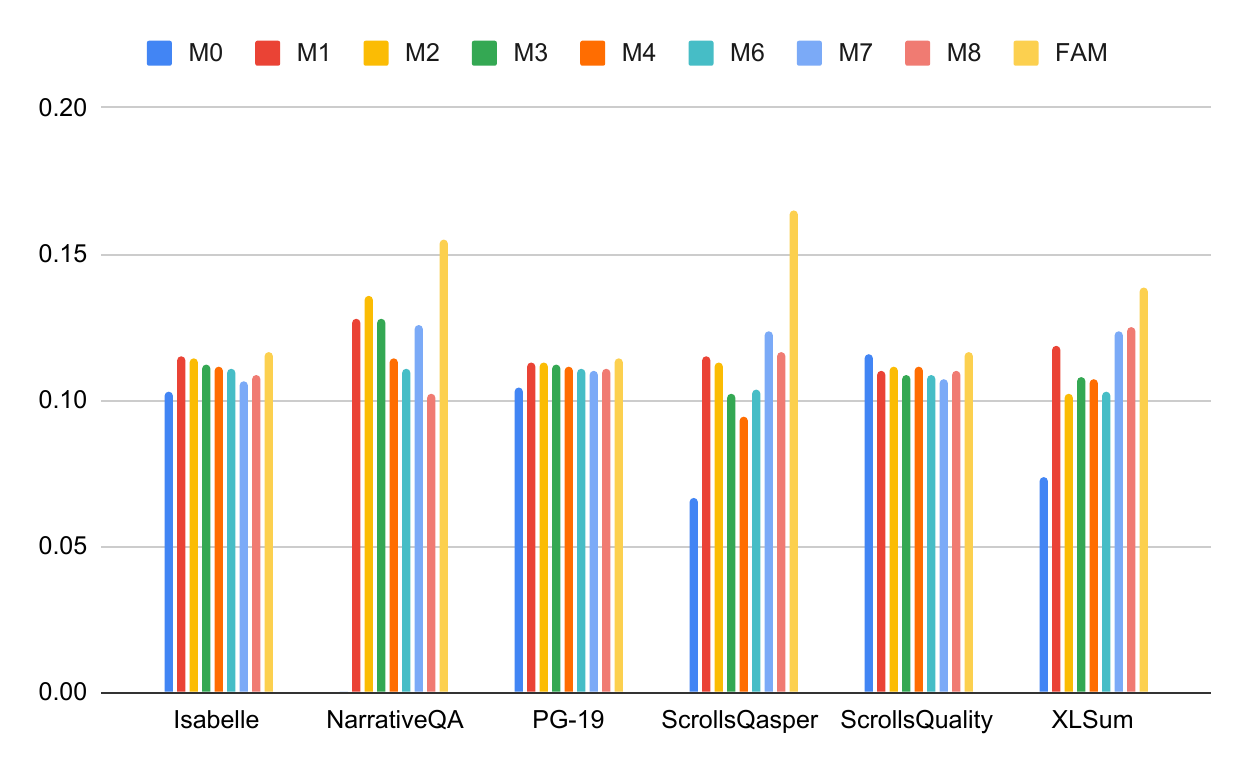}
    \caption{Long Context Tasks \label{fig:lct}}
    \end{subfigure}\hfill
    \caption{ (a) PassKey Retrieval: Performance across different Transformer models and memory segment configurations. MX denotes the number of BSWA memory segments. FAM represents TransformerFAM with 0 memory segments. TransformerFAM successfully solves the task. (b) LCT: Normalized scores of long-context tasks evaluated by Flan 1B with different Transformer models and different memory segment configurations. FAM outperforms all other BSWA configurations.}
    % \vspace{-4mm}
\end{figure*}

As shown in \cref{fig:passkey}, \textbf{TransformerFAM was able to perfectly solve the task with a filler context of up to 260k tokens}. In the figure, MX denotes the number of BSWA memory segments. The performance of TransformerBSWA improves significantly up to M2, after which it saturates. The performance of M12 also drops significantly after 20k tokens. The theoretical receptive field of M2 is 36k (i.e. depth(18) $ \times $ memory segment(2k)), but the effective receptive field is much shorter.

% \begin{figure}[ht]
% % \vskip 0.2in
% \begin{center}
% \centerline{\includegraphics[width=\columnwidth]{fig/passkey}}
% \caption{ PassKey Retrieval: Performance across different Transformer models and memory segment configurations. MX denotes the number of BSWA memory segments. FAM represents TransformerFAM with 0 memory segments. TransformerFAM successfully solves the task.}
% \label{fig:passkey}
% \end{center}
% % \vskip -0.2in
% \end{figure}

In \cref{fig:passkey}, it is important to compare M1\_SG and M1. M1\_SG has a stop gradient applied to one memory segment, which limits the receptive field to the window size. This is because the model cannot learn which contextual information stored in the memory segment will have a good result later. SWA with back propagation through time (BPTT) functions similarly to a time-limited RNN. It is common to use stop gradients on the memory segments of TransformerXL~\cite{dai2019transformer,chevalier2023adapting}. However, we recommend against this practice.

In \cref{sssec:rmt}, we compare our work with recent Transformer with memory papers~\cite{bulatov2022recurrent,chevalier2023adapting}.

\subsection{Long Context Tasks}
\label{ssec:lct}

Gemini~\cite{team2023gemini} evaluated long-context capabilities using the following tasks: NarrativeQA~\cite{kovcisky2018narrativeqa}, Scrolls-Qasper, Scrolls-Quality~\cite{shaham2022scrolls}, and XLSum~\cite{hasan2021xl}. Additionally, PG-19~\cite{rae2019compressive} and Isabelle~\cite{wu2022memorizing} are another common evaluation tasks among long-context Transformer papers~\cite{rae2019compressive,wu2022memorizing,chen2023extending}. Detailed information on the evaluation data is provided in \cref{table:lct} in \cref{app:lct}.

% \begin{figure}[ht]
% % \vskip 0.2in
% \begin{center}
% \centerline{\includegraphics[width=\columnwidth]{fig/lct}}
% \caption{ Normalized scores of long-context tasks evaluated by Flan 1B with different Transformer models and different memory segment configurations. FAM outperforms all other BSWA configurations.}
% \label{fig:lct}
% \end{center}
% % \vskip -0.2in
% \end{figure}

We evaluated the long-context capabilities of the 1B TransformerBSWA model trained in \cref{ssec:training} using memory segment sizes ranging from 0 to 8. As shown in \cref{fig:lct}, TransformerFAM outperformed TransformerBSWA on all the long context tasks (LCT), regardless of the number of memory segments in BSWA. It shows a significant performance improvement on ScrollsQasper and NarrativeQA, where it has to understand 5k to 500k tokens of context before answering a question. \textbf{The LCT results demonstrate that TransformerFAM can effectively compress and retain important contextual information within extremely long contexts}. 

Above M1, the number of memory segments does not significantly impact LCT performance on TransformerBSWA, because the input sequences are much longer than the window size of all experiments. We observed the same phenomenon in TransformerFAM, and TransformerFAM uses \mseg memory segments in \cref{fig:lct}. The figure shows the normalized scores of all tasks to view the scores on the same scale. The raw results are in \cref{table:lct_results} in \cref{app:lct}.

We further evaluated TransformerFAM and TransformerBSWA on 8B and 24B models. As shown in \cref{table:8b_lct}, TransformerFAM demonstrates scalability as the model size increases. This suggests that self-attention can route local information relevant to each input sequence while simultaneously routing contextual information to the FAM.
However, the performance improvements are not substantial, indicating room for further enhancements in working memory mechanisms.

\begin{table}[h]
    % \small
    \centering
    \begin{tabular}{lcccc}
        \toprule
        \bf Model & \bf \makecell[c]{BSWA \\ 8B} & \bf \makecell[c]{FAM \\ 8B} & \bf \makecell[c]{BSWA \\ 24B} & \bf \makecell[c]{FAM \\ 24B} \\
        \midrule
        Isabelle & 82.1 & \bf82.5 & 86.6 & 86.6 \\
        NarrativeQA & 18.4 & \bf19.3 & 22.6 & \bf23.0 \\
        PG-19 & 52.4 & \bf52.9 & 55.7 & \bf57.2 \\
        ScrollsQasper & 12.4 & \bf18.5 & 28.0 & \bf29.4 \\
        ScrollsQuality & 47.3 & \bf48.5 & 55.4 & \bf58.0 \\
        XLSum & 22.0 & \bf24.7 & 24.7 & \bf26.4 \\
        \bottomrule
    \end{tabular}
    \caption{
        LCT scores on 8B and 24B models comparing TransformerBSWA and TransformerFAM
    }
    \label{table:8b_lct}
\end{table}

In addition, TransformerFAM marginally surpasses TransformerBSWA on GPT-3 tasks~\cite{brown2020language} (see Table~\ref{table:gpt3}). 
This result is unexpected since all tasks involve sequences shorter than 2k tokens. We hypothesize that this improvement arises from the efficient contextual representation  by TransformerFAM. \textbf{By offloading contextual data to FAM, TransformerFAM reduces redundancy within input activations, optimizing latent space usage}.\footnote{It can be viewed as the decoder analog of register tokens~\cite{darcet2023vision} in ViT encoders~\cite{dosovitskiy2020image} to process global context.}

\begin{table}[h]
    % \small
    \centering
    \begin{tabular}{lcc}
        \toprule
        \bf Model & \bf GPT-3 Rank & \bf GPT-3 Gen \\
        \midrule
        BSWA 1B  & 60.2 & 33.9 \\
        FAM 1B  & \textbf{61.0} & \textbf{34.7} \\
        BSWA 8B & 72.8 & 54.3 \\
        FAM 8B & \textbf{74.0} & \textbf{54.9} \\
        BSWA 24B & 78.2 & 62.6 \\
        FAM 24B & \textbf{78.5} & \textbf{63.4} \\
        \bottomrule
    \end{tabular}
    \caption{
        Summarizes GPT-3 performance on ranking and generative tasks. (Details in \cref{table:gpt3_raw}) % in \cref{app:gpt3}.
    }
    \label{table:gpt3}
\end{table}

Thus, BSWA memory segments (local representation) and FAM (global representation) complement each other. For LLMs, we recommend using FAM for compressed contextual representation alongside BSWA memory segments up to inference budgets (e.g., 2k, 8k, 32k~\cite{team2023gemini}, or 128k~\cite{achiam2023gpt}).
Due to page limitations in the main paper, ablation studies are presented in \cref{ssec:ablation}.

% \subsection{Ablation study}

% FIXME: Remove it and move Appendix here for arxiv paper.
% We conducted ablation studies on each hyperparameter.

% \weiran{I suggest saving space as mentioned in previous sections, and present more ablation studies.} <- let me try

%% file: _2-related.tex
\section{Related Work}
\label{related}

There have been attempts to incorporate feedback mechanisms into the Transformer, but most of them involve feeding the output activations from the top layer to the bottom~\cite{bulatov2022recurrent,chevalier2023adapting} or to intermediate layers~\cite{fan2020addressing}. Since the top three layers in the Transformer are heavily focused on output reconstruction~\cite{pasad2021layer}, we hypothesize that there is a significant representational gap between the top and other layers. In this paper, we propose a feedback mechanism between intermediate layers.

There were papers that compressed information blockwise~\cite{rae2019compressive,guo2019star,gupta2020gmat,mohtashami2023landmark,mu2023learning}. However, in those papers, the information was not propagated infinitely. Relevant prior work includes the use of recurrent cross-attention between blocks~\cite{hutchins2022block}, enabling the propagation of compressed information to subsequent blocks. Additionally, incorporating feedback from a few upper layers has been used to integrate past information~\cite{ju2022staircase}. We propose TransformerFAM under the assumption that the human brain processes homogenous, heterogeneous, and feedback data with the same attention mechanism across distributed brain areas.
Additional related works are presented in \cref{app:related}.
% Due to page limitations in the main paper, additional related works are presented in \cref{app:related}.

%% file: _5-conclusion.tex
\section{Conclusion}
\label{conclusion}

In the film 'Memento' (2000), the protagonist struggles with anterograde amnesia, which means he can not remember anything before happened in the last 10 minutes, but his long-term memory is intact, He has to tattoo important information on his body to remember it. This is similar to the current state of large language models (LLMs). LLMs memorize the entire internet thanks to scaling laws~\cite{kaplan2020scaling}, which allow them to store an enormous amount of information in large weights (long-term memory). However, their short-term memory is limited by the attention window. As a result, the complex prompt engineering becomes necessary to help them recall important details. We propose a new architecture called TransformerFAM that could fix anterograde amnesia of LLMs.

The rapid progress of machine learning is astonishing, but there are two key problems that we still do not know how to approach: reasoning and memory. In this paper, we provide a clue to the memory problem. Memory is a critical prerequisite for reasoning. It is hard to imagine how we can derive complex mathematical equations without working memory. Reasoning must be a phenomenon that occurs based on the current working memory.

This paper explores the integration of attention-based working memory, a concept from neuroscience, into the field of deep learning. Our goal is to ignite further research within the community to address and solve the ongoing challenge of limited memory in deep learning. There is a significant set of problems to tackle here, ranging from refining feedback attention architecture to investigating the transfer of working memory to long-term memory.

%% file: _9-appendix.tex
\newpage
\appendix
\onecolumn

\section{Architecture details}
\label{app:arch}

\subsection{Flan-PaLM architecture}
\label{app:flan}

\cref{table:llm_arch} provides detailed information on the architecture of Flan-PaLM for 1B, 8B, and 24B models. MQA (Multi-Query Attention)~\cite{shazeer2019fast} is an attention mechanism that employs a single set of keys and values for all attention heads.

\begin{table}[h]
    \centering
    
    \begin{tabular}{lccc}
    \toprule
    \makecell[c]{Component} & Flan-PaLM 1B & Flan-PaLM 8B & Flan-PaLM 24B \\
    \midrule
    Num. Layers   & 18 & 32 & 56 \\
    Model Dim.    & 1536 & 4096 & 4096 \\
    FF Multiplier & 8 & 4 & 8 \\
    Num. Heads    & 12 & 32 & 32 \\
    Use MQA       & F & F & T \\
    \bottomrule
    \end{tabular}
\caption{ Architecture of 1B, 8B, and 24B Flan-PaLM models}
\label{table:llm_arch}
\end{table}

\subsection{FAM hyperparameters}
\label{app:fam_hparam}

\cref{table:fam_hparam} presents the default settings for the hyperparameters added in TransformerFAM.

\begin{table}[h]
    
    \centering
    \begin{tabular}{lc}
        \toprule
        \bf Component & \bf Value \\
        \midrule
        Memory Segment & \mseg \\
        FAM length & \flen \\
        Probability of Random State Passing & 0.8 \\
        \bottomrule
    \end{tabular}
    \caption{
        TransformerFAM hyperparameters
    }
    \label{table:fam_hparam}
\end{table}

\section{Additional details of TransformerFAM}
\label{app:additional_fam}

The appendix describes additional details not covered in the main text.

\subsection{FAM initialization}
\label{ssec:init_feedback}

The current FAM $F_\tau$ is the output activation of the previous FAM update. So, how do we initialize the FAM for the first block?

When FAM are used as queries, they summarize the block context. Therefore, we learn by adding learnable summarization embeddings to the token embedding lookup level of the Transformer model. This is the same as prepending learnable embeddings in soft prompt tuning~\cite{lester2021power}. The difference is that full attention is applied between the FAM prompt activations, and the updated FAM is used for the next block. The FAM prompt is passed to the next transformer layer through the forward propagation of the Transformer model, and it has a summary representation that is suitable for the level of the layer.

Prefix tuning~\cite{li2021prefix} can also be used to train learnable initial FAM at each Transformer layer. However, we had difficulty matching the magnitude of the learnable prefix to the regular input sequence, and the results of prompt tuning were consistently better. Ablation study in \cref{sssec:prefix} shows prompt tuning outperforms. In addition, prefix tuning has the disadvantage of adding additional weights of FAM length to each layer.

In addition, the first FAM update in self-attention should utilize a zero tensor in the residual connection rather than the initial FAM, because the initial FAM does not carry any contextual information.

In summary, we learned the initial FAM using prompt tuning, which only adds a very small number of weights of FAM length to the entire model. We share the same initial FAM across all minibatches.

\subsection{Input Length Extrapolation}
\label{ssec:position}

\subsubsection{FAM Position Encoding}
\label{sssec:pos}

We used rotary position embedding (RoPE)~\cite{su2024roformer} in all of our experiments. Each input sequence is assigned an integer position $m$, which is converted into sinusoidal encoding in the form of $\exp(i m \theta)$ where $\theta = 10000^{-2i / d_{model}}$~\cite{vaswani2017attention}.

FAM is inserted at each block boundary, but the problem is how to assign positions to FAM. We assigned positions to FAM in order from the last number of the compressed block. For example, if the block size is 4, the FAM length is 2, and the positions of the compressed blocks are $m$, $m+1$, $m+2$, and $m+3$, then the updated FAM positions are $m+2$ and $m+3$.

We tried other methods, but this method worked best according to the ablation results in \cref{sssec:fam_pos}.

\subsubsection{Random Position Offset}
\label{sssec:rpo}

The input length extrapolation problem of Transformer is well known~\cite{chen2023extending,xiong2023effective}. For example, a Transformer LLM trained with 2k tokens experiences a severe performance drop when generating 8k tokens. This is because machine learning (ML) does not generalize well to situations that it has not seen during training.

Transformer with SWA does not suffer from the position extrapolation problem when using relative positional embedding like RoPE. This is because the score value of $q_m \cdot k_n$ becomes a function of $ (m - n) $ in the form of $\exp(i (m - n) \theta)$.

Because the range of $ (m - n) $ is limited to the window size, independent of the input length, the model can handle long input sequences without facing novel scenarios during inference. The model can accurately determine the relative position, if the window size is smaller than the maximum wavelength of $\theta$. If the typical $\theta$ is used for positional embedding, the working maximum window size is the maximum wavelength ($= 2 \times 10000 \times \pi \sim 63$k tokens ).

However, FAM breaks the symmetry of relative position. Since the absolute position from the past to the present is recursively embedded in the FAM, the large absolute position value that the model encounters for the first time during inference creates a situation where the model needs to extrapolate.

We propose Random Position Offset as a solution. At each training step, the Transformer model randomly samples a scalar value between 0 and the maximum wavelength. All Transformer layers add that random value to the absolute position at that training step. Therefore, the FAM experiences the entire range of absolute position embedding during training.

This is a purely training technique. During inference, the default offset is $0$. We used the below algorithm that generates $0$ by $50\%$ when sampling the offset, as $0$ is the default value.

\begin{lstlisting}[language=Python]
offset = np.uniform([b], maxval=wavelen)
offset *= np.round(np.uniform([b]))
\end{lstlisting}

\subsubsection{Random State Passing}
\label{sssec:rsp}

Due to the recursive use of FAM, we need to determine the maximum number of updates for which the FAM remains valid. If it is updated up to 8 times during training, the model will have to extrapolate the situation where it is updated 100 times during inference.

The same problem existed in RNNs, and Random State Passing (RSP)~\cite{narayanan2019recognizing} was proposed to handle long-form speech in the speech domain. We also used RSP to generalize the number of updates of FAM.

RSP saves FAM as weights at the end of each training step. Then, it loads the saved FAM at the next training step. When initializing FAM, it either uses randomly saved FAM or learned FAM. In our default setup, it used saved FAM with $80\%$ probability. To save FAM of all minibatch, weights are required as many as the number of minibatch. We save FAM of only the first batch and all minibatch share it in the next training step.

On the other hand, saved FAM can be thought of as learnable prefix for prefix tuning~\cite{li2021prefix}. It is also possible to train only the FAM while the model is frozen and use them for various downstream tasks or personalization. This part is left as a future research topic.

\subsection{FAM illustrated}
\label{ssec:fam_illustrated}

\begin{figure*}[ht]
    \centering
    \begin{subfigure}{.3\textwidth}
    \includegraphics[width=\linewidth]{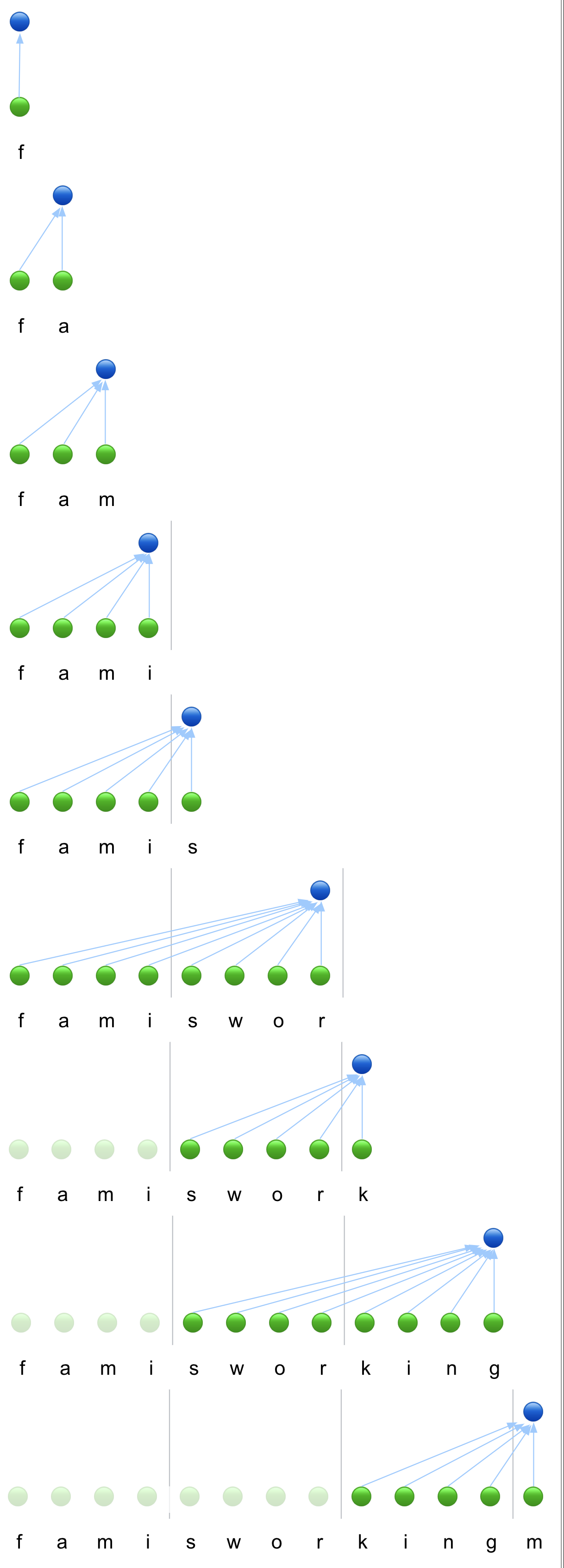}
    \caption*{TransformerBSWA}
    \end{subfigure}\hfill
    \begin{subfigure}{.19\textwidth}
    \includegraphics[width=\linewidth]{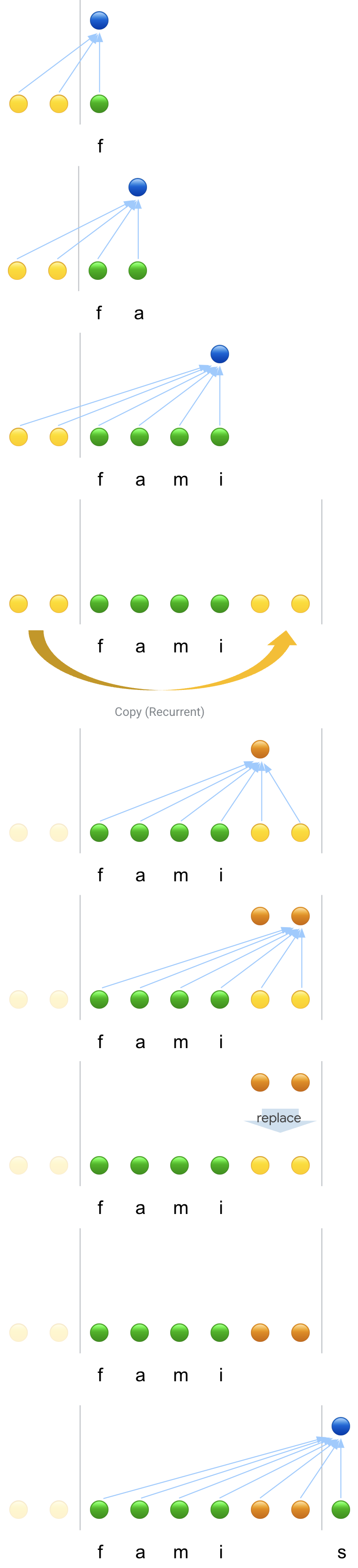}
    \caption*{}
    \end{subfigure}\hfill
    \begin{subfigure}{.48\textwidth}
    \includegraphics[width=\linewidth]{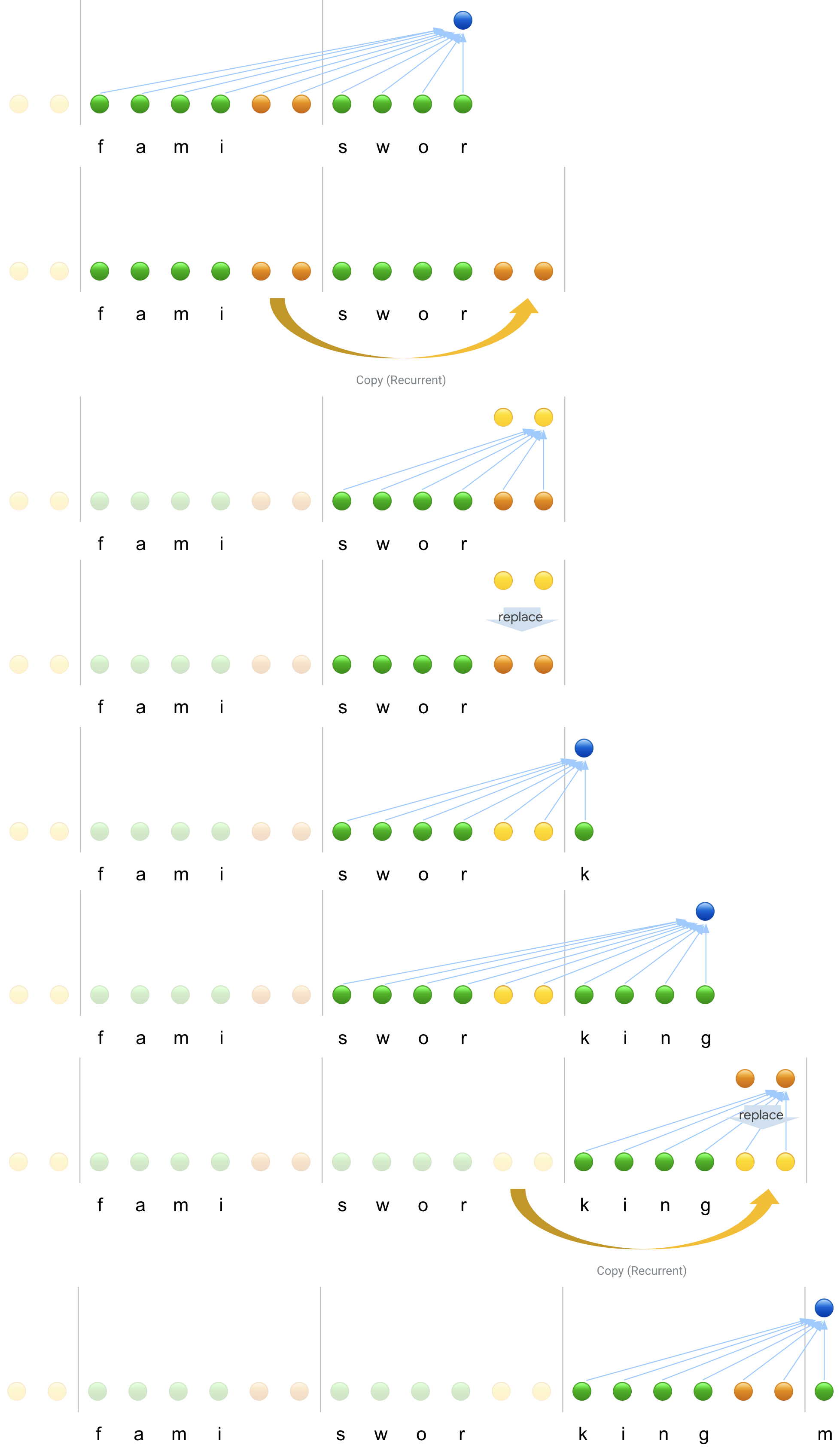}
    \captionsetup{singlelinecheck=false}
    \caption*{TransformerFAM}
    \end{subfigure}\hfill
    \caption{ Visualization of self-attention during inference over time. (A) Self-attention pattern of TransformerBSWA layer with memory segment size of 1. (B) Self-attention pattern with FAM added.  \label{fig:animation}}
    % \vspace{-4mm}
\end{figure*}

In \cref{ssec:fam}, we define the feedback loop as feedback activations that feed contextual representation back into each block of BSWA. This feedback loop is formally described in \cref{alg:fam}. While \cref{fig:transformer_fam} illustrates \cref{alg:fam}, the static image makes it challenging to fully grasp the dynamic nature of the decoding self-attention mechanism. To clarify this, we create a multi-frame animation in \cref{fig:animation} that demonstrates how the attention mechanism evolves over time (top to bottom).

% https://docs.google.com/drawings/d/1MTPYUXFfpw0kqTX9lrHqajD_WJ2OxGUDy4Ek3psEKmA/edit
% https://docs.google.com/drawings/d/10mwSQcUX1SdjI-IIc6OhDCXpKy4bofHZbVq0IBDAP3w/edit
% https://docs.google.com/drawings/d/13BoXCqhUT0Zc2XBaxG1pvvdTnZldw5sT8urcbCBwud8/edit
% https://docs.google.com/presentation/d/1J9Zs9Ql0d-vmSlslwbBnRQJj6wMiGFlK59Sb-4fY2Bc/edit

\subsection{Ablation Studies}
\label{ssec:ablation}

All ablation studies were conducted on a 1B model.

\subsubsection{FAM length}
\label{sssec:abl_fam_len}

In the Flan 1B model, we observed performance saturation on ScrollsQasper, ScrollsQuality, and XLSum tasks when the FAM length reached 64. Interestingly, performance declined when FAM length exceeded 64, suggesting that information compression is more effective with limited space. This constraint on memory capacity is reminiscent of Miller's Law~\cite{miller1956magical}, which posits that the average person can only hold approximately $7 (\pm 2)$ items in their working memory at any given time.
% The limited space is a feature, not a bug.

\begin{table}[h]
    % \small
    \centering
    \begin{tabular}{lccc}
        \toprule
        \bf FAM Len. & \bf ScrollsQasper  & \bf ScrollsQuality  & \bf XLSum \\
        \midrule
        4 & 5.0 & 27.1 & 15.1 \\
        16 & 6.0 & 25.2 & 15.2 \\
        64 & \bf 7.2 & \bf 27.9 & 15.9 \\
        256 & 5.1 & 26.5 & \bf 16.0 \\
        1024 & 5.3 & 26.3 & \bf 16.0 \\
        \bottomrule
    \end{tabular}
    \caption{
        LCT scores according to FAM Length
    }
    \label{table:fam_len}
\end{table}

\subsubsection{The number of previous FAM blocks}
\label{sssec:num_fam}

In \cref{fig:transformer_fam}, the input query attends to the FAM as denoted Attention to Feedback. The input query can attend to not only the immediately previous FAM, but also to more previous FAMs. \cref{table:fam_blocks} shows the XLSum scores for different numbers of previous FAM blocks. As the table shows, increasing the number of blocks did not have a significant effect, because the previous FAM already encapsulates all the previous information by a feedback copy. Therefore, the default setup attends to only the immediately previous FAM.

\begin{table}[h]
    % \small
    \centering
    \begin{tabular}{lccccc}
        \toprule
        \bf FAM blocks & 1 & 2 & 3 & 4 & 6 \\
        \midrule
        \bf XLSum & 15.9 & 15.8 & 15.7 & 15.1 & 16.2 \\
        \bottomrule
    \end{tabular}
    \caption{
        XLSum scores according to the number of FAM blocks
    }
    \label{table:fam_blocks}
\end{table}

\subsubsection{FAM Position Encoding}
\label{sssec:fam_pos}

\cref{sssec:pos} proposed assigning the last number of the compressed block as the FAM position. In addition, we also experimented with FAM having a float position between blocks. In the example of block size 4 in \cref{sssec:pos}, the FAM positions would be $m+3+0.33$ and $m+3+0.66$. We also experimented with the case where the FAM position is always 0.

As shown in \cref{table:fam_pos}, the last number showed the best accuracy.

\begin{table}[h]
    
    \centering
    \begin{tabular}{lcc}
        \toprule
        \bf FAM position & \bf PG-19 & \bf Isabelle \\
        \midrule
        \bf Last number & \bf 47.7 & \bf 73.6 \\
        Float number & 47.6 & 73.4 \\
        Zero & 47.3 & 72.1 \\
        \bottomrule
    \end{tabular}
    \caption{
        PG-19 and Isabelle Accuracy across various FAM Position Encoding
    }
    \label{table:fam_pos}
\end{table}

\cref{fig:pg19} shows FAM (i.e. Last number in \cref{table:fam_pos}) outperforming FAM-POS (i.e. Zero) in PG-19 accuracy over most base frequencies.

\subsubsection{Random Position Offset}

As mentioned in Section~\ref{sssec:rpo}, the input length extrapolation problem of the Transformer is well-known. The "Attention is All You Need"~\cite{vaswani2017attention} introduced sinusoidal position encoding in the form of $\exp(i (m - n) \theta)$ where $\theta = 10000^{-2i / d_{model}}$. Popular solutions for full attention models include increasing the base frequency from 10k to 500k~\cite{chen2023extending} or scaling down $ (m - n) $ to $ (m - n) \times (1024 / 4096) $~\cite{xiong2023effective}.

Since the relative position of BSWA has a range of 0 to window size, it does not suffer from the input length extrapolation problem. However, TransformerFAM encodes absolute position into FAM, which requires a solution. Section~\ref{sssec:rpo} proposes Random Position Offset (RPO) as a solution.

\begin{figure}[h]
% \vskip 0.2in
\begin{center}
\centerline{\includegraphics[width=0.6\textwidth]{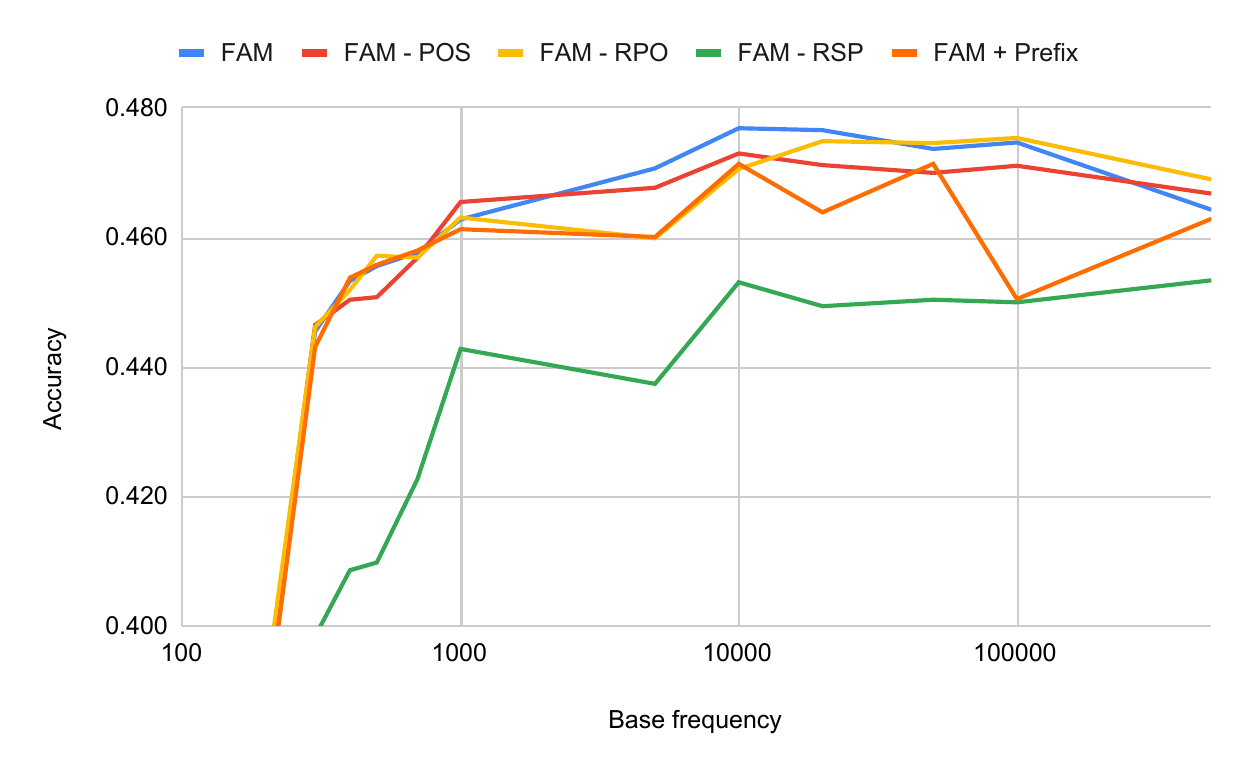}}
\caption{ PG-19 accuracy for various ablation studies such as RPO (Random Position Offset), RSP (Random State Passing) and Prefix FAM tuning over different base frequency.}
\label{fig:pg19}
\end{center}
% \vskip -0.2in
\end{figure}

In \cref{fig:pg19}, FAM shows better PG-19 accuracy than FAM-RPO at the 10k base frequency. As mentioned in \cref{app:lct}, the max length of PG-19 was truncated to 256k in the experiments.

Furthermore, scaling up the base frequency or scaling down $ (m - n) $ is only a remedy for pain, not a final solution. It reduces the resolution of the position, which negatively affects the overall Transformer performance, as shown in the figure. Interestingly, the originally proposed 10k is a very good value even in long contexts, until the window size reaches its wavelength (63k).

In addition, we did not observe the attention sink phenomenon~\cite{xiao2023efficient} in our TransformerBSWA experiments. The paper proposes that Transformers with sliding window attention should store initial tokens in a KV cache for long-context inference. However, the relative position of SWA is restricted to a range of 0 to window size, independent of input sequence length. Our TransformerBSWA implementation, trained on 8.5k tokens, successfully operated for inference up to 256k tokens without any issues.

\subsubsection{Random State Passing}

To extrapolate the number of FAM updates, \cref{sssec:rsp} proposes Random State Passing (RSP). In \cref{fig:pg19}, FAM shows significantly better PG-19 accuracy than FAM-RSP. This demonstrates that RSP plays a crucial role in training FAM.

\begin{figure}[h]
% \vskip 0.2in
\begin{center}
\centerline{\includegraphics[width=0.6\textwidth]{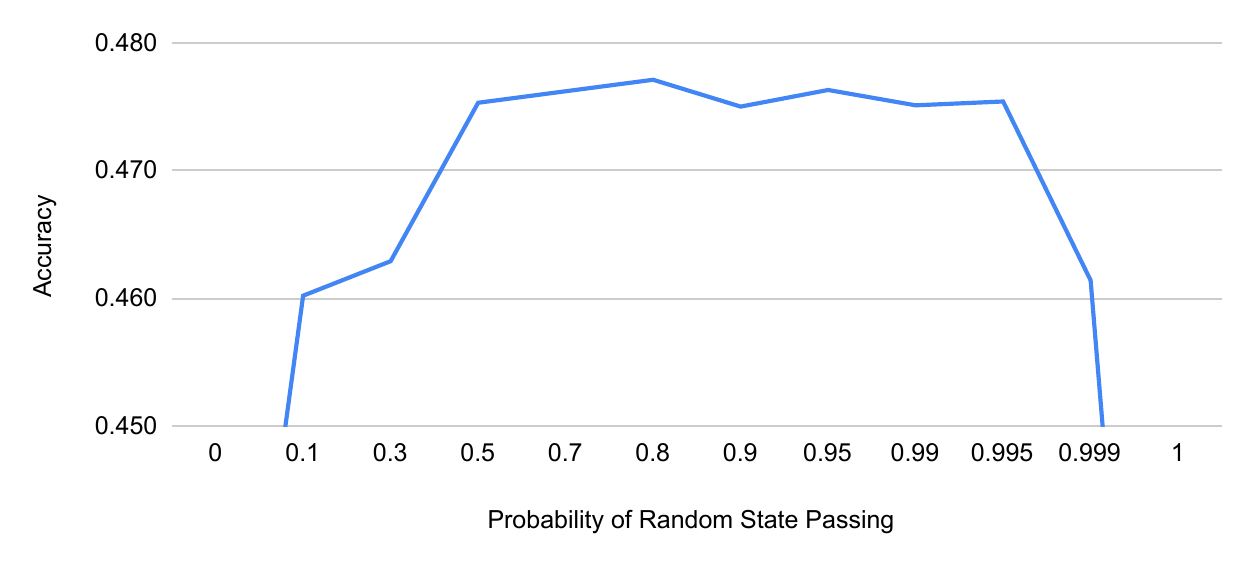}}
\caption{ PG-19 accuracy for different probability of Random State Passing.}
\label{fig:rsp}
\end{center}
% \vskip -0.2in
\end{figure}

\cref{fig:rsp} shows the best performance at a probability of 0.8. At 0.8, the half-life is 3 training steps ($ 0.512 = 0.8^3 $). This means that every 26k (8.5k x 3) tokens, FAM restarts from the beginning with a 50$\%$ probability. In other words, if FAM experiences 25 FAM updates during training, it can extrapolate to 256 FAM updates (256k tokens) during inference.

\subsubsection{Prompt vs. Prefix}
\label{sssec:prefix}

\cref{ssec:init_feedback} mentions that Prompt tuning outperforms Prefix tuning for training FAM in terms of performance, memory, and number of parameters. \cref{fig:pg19} shows that Prefix tuning for FAM training leads to performance degradation in PG-19 (FAM vs. FAM+Prefix).

\subsubsection{TransformerXL vs. TransformerBSWA}
\label{sssec:txl}

TransformerXL is a modification of TransformerBSWA that incorporates an additional QK attention mask to mask out keys beyond a specified window size. Therefore, there is no reason to expect a significant difference in performance between the two. \cref{table:txl} shows the scores for the main tasks. TransformerBSWA used a single memory segment, and TransformerXL window size was equal to the block size of \block.

\begin{table}[h]
    
    \centering
    \begin{tabular}{lcc}
        \toprule
        \bf Tasks & \bf TransformerBSWA & \bf TransformerXL \\
        \midrule
        Isabelle & 72.6 & 72.2 \\
        NarrativeQA (RougeL) & 11.1 & 11.0 \\
        PG-19 & 46.4 & 46.1 \\
        ScrollsQasper (RougeL) & 5.0 & 4.9 \\
        ScrollsQuality (accuracy) & 26.3 & 26.5 \\
        XLSum (RougeL) & 13.6 & 13.7 \\
        GPT-3 Rank & 60.2 & 59.9 \\
        GPT-3 Gen & 33.9 & 33.9 \\
        \bottomrule
    \end{tabular}
    \caption{
        Comparing TransformerBSWA and TransformerXL on major tasks
    }
    \label{table:txl}
\end{table}

\subsubsection{Comparison with other methods}
\label{sssec:rmt}

\begin{figure}[h]
% \vskip 0.2in
\begin{center}
\centerline{\includegraphics[width=0.6\textwidth]{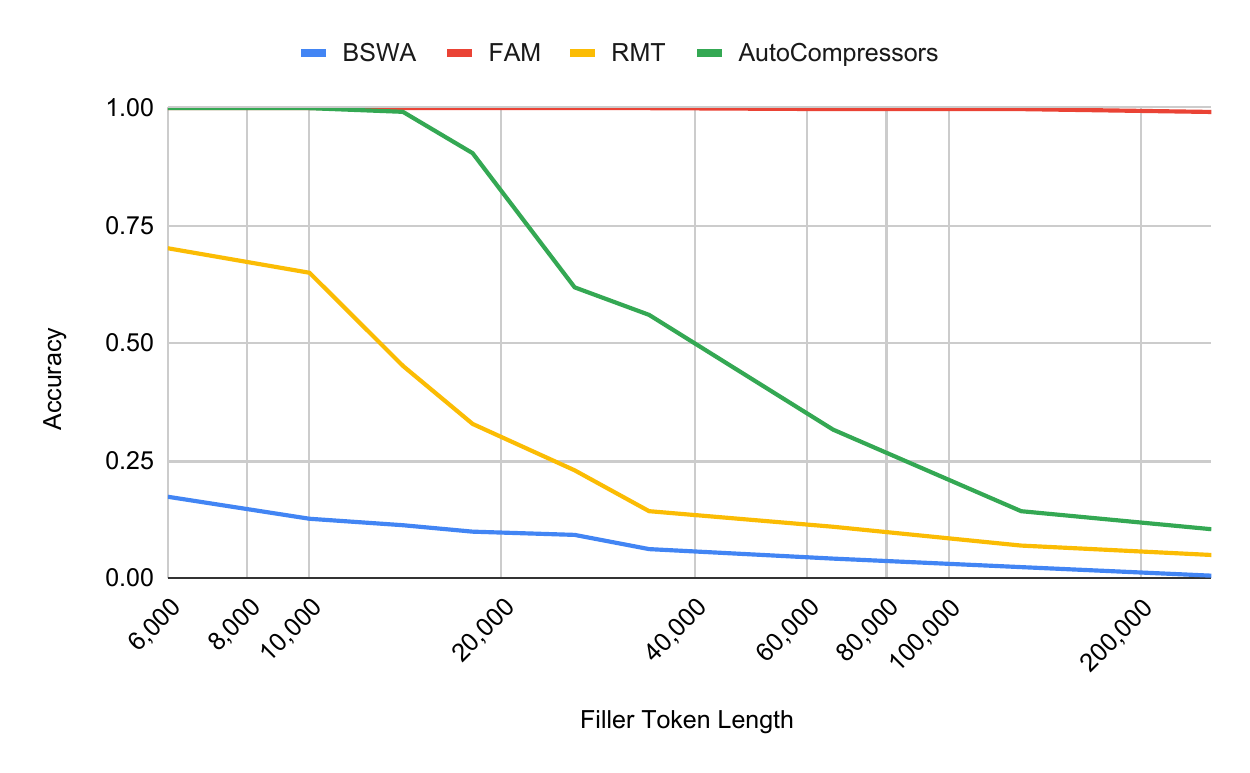}}
\caption{ PassKey Accuracy: FAM maintains performance with long sequences, outperforming BSWA, RMT and AutoCompressors. }
\label{fig:rmt}
\end{center}
% \vskip -0.2in
\end{figure}

We compared FAM with Recurrent Memory Transformer (RMT)~\cite{bulatov2022recurrent}, because RMT also implements memory in Transformer using feedback mechanism from top layer to bottom layer. 
As shown in \cref{fig:rmt}, RMT showed worse performance than Block Sliding Window Attention (BSWA) with 1 memory segment, in the PassKey retrieval task. FAM solved the PassKey retrieval task, but RMT did not work at all with very long filler token lengths. 

RMT is implemented by feeding the output memory of the previous segment as the input memory to the next segment. In the constraint that the input is text embedding and the output is text reconstruction in LLM, RMT has an additional constraint that the latent space of the output memory and the input memory must match. In this situation, RMT fails to remember the PassKey for a very long context. On the other hand, FAM seems to compress, store, and propagate information more effectively by learning memory representation that matches the abstraction level of each layer through training.

AutoCompressors~\cite{chevalier2023adapting} is an extension of RMT that continuously accumulates blockwise memory of RMT. AutoCompressors theoretically should be able to solve the PassKey task since it maintains all the memory tokens for all blocks. However, as shown in \cref{fig:rmt}, its performance drops sharply after 18k tokens. This is because the model only saw up to 18k tokens during training. It fails to generalize to longer filler token lengths. The AutoCompressors in \cref{fig:rmt} accumulates 260 memories to support up to 260k tokens.

\section{Don't}
\label{app:dont}

There could be various ways to create feedback loops in Transformers. This appendix summarizes our attempts that did not work well. We hope that this will save other researchers time when improving the architecture of feedback loops in future studies.

\subsection{Beyond Block-Recurrent Transformers}

Block-Recurrent Transformers (BRT)~\cite{hutchins2022block} have a recurrent state that connects each block, and the recurrent state and the block input sequence exchange information through cross-attention. In this architecture, the recurrent state plays a role similar to FAM. We started the project by removing the additional QKV projection and cross-attention for the recurrent state in BRT and integrating all computations into self-attention.

Like BRT, we tried to compress the block input by attention to use it as the recurrent state of the next block, and it required additional projections like BRT. It was difficult to properly train the additional projections that were only used when striding blocks, and as a result, the performance of general tasks such as GPT-3 tasks was degraded. Finally, we found that the activations compressed by attention must go through the FF layer of the Transformer to be aligned as a latent representation that can be used again as the input of the next Transformer. The input of the Transformer and the output of self-attention are very different representations. However, the FF layer transforms it back into a representation similar to the input of the Transformer for the next Transformer layer. After the discovery of reusing the FF layer, we also found that separate QKV projection and additional projections are not required for FAM. That is how the TransformerFAM architecture was created.

Around the time this paper was published, Infini-Transformer~\cite{munkhdalai2024leave} was also released. Infini-Transformer reported successful PassKey retrieval despite the compressed information not passing through the FF layer. This is because it stores weighted values in scratchpad memory and reuses them, preserving the latent space of the values. However, since the weighted values are all it can represent, there might be limitations in its expressiveness.

Furthermore, TransformerFAM maintains the past memory segment from BSWA despite having compressed memory because the compressed memory cannot retain detailed information. Since Infini-Transformer completely discards the past memory segment, it might have difficulty remembering recent details.

\subsection{Feedback Memory Segment}

\begin{figure}[h]
% \vskip 0.2in
\begin{center}
\centerline{\includegraphics[width=0.7\textwidth]{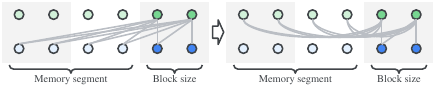}}
\caption{ Convert TransformerBSWA to TransformerFM }
\label{fig:xformer_feedback}
\end{center}
% \vskip -0.2in
\end{figure}

As shown in \cref{fig:transformer_fam}, TransformerFAM utilizes the updated FAM from the previous block as the input to the current block. Therefore, it is natural to consider using the Transformer outputs of the previous block directly as the input to the current block, instead of using a complex FAM mechanism. \cref{fig:xformer_feedback} illustrates this modification, which we refer to as the Feedback Memory Segment (FM). FM achieves an infinite theoretical receptive field by performing self-attention at the output level. This is similar to ERNIE-Doc~\cite{ding2020ernie}, and a specific variant of Staircase Attention~\cite{ju2022staircase}. Staircase Attention proposes using activations from progressively higher Transformer layers as the memory segment goes further into the past.

% https://docs.google.com/drawings/d/1av2XYdFbhV0JA8gC-QnODWHEV9MqCY4-aAX62Vz22aM/edit?resourcekey=0-dplOiwAH8SuL_954lXY4sQ

\begin{figure}[h]
% \vskip 0.2in
\begin{center}
\centerline{\includegraphics[width=0.6\textwidth]{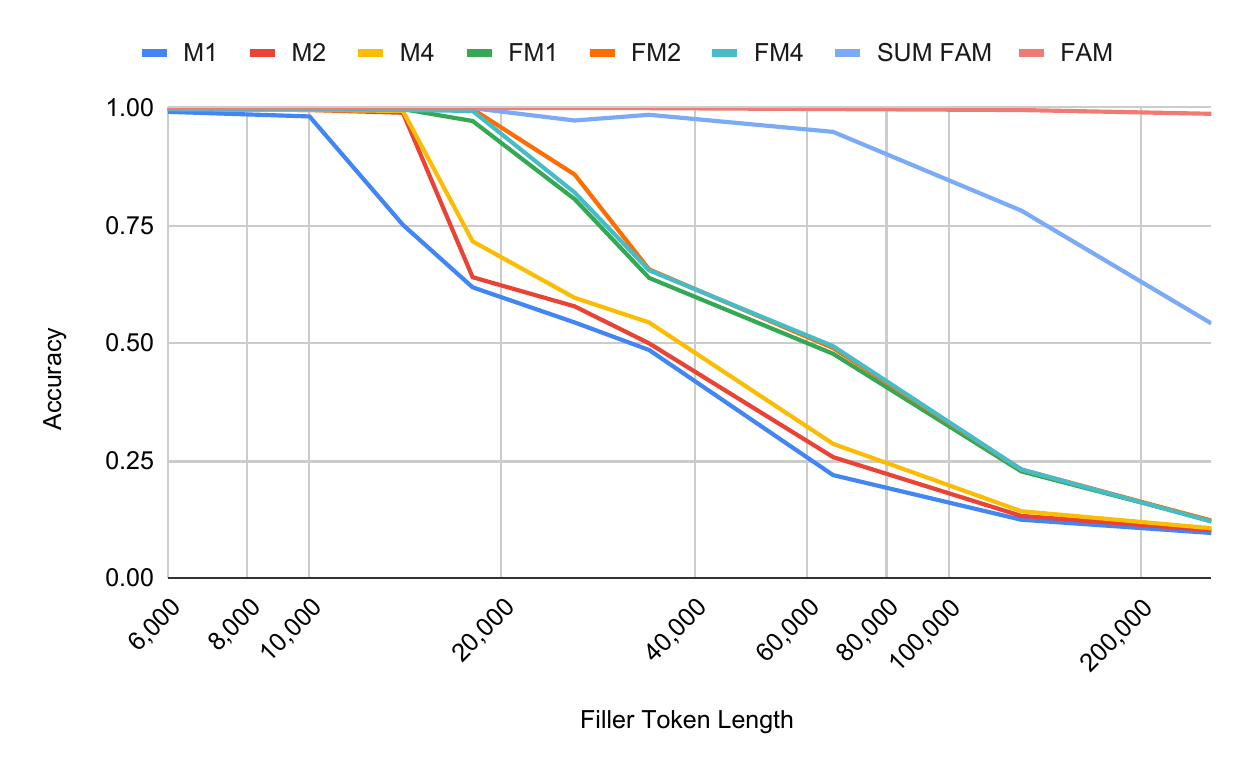}}
\caption{ PassKey Retrieval Accuracy Using BSWA (M1,2,4), Feedback Memory Segments (FM1,2,4), Static Summary Tokens (SUM FAM) and FAM}
\label{fig:sum_fms}
\end{center}
% \vskip -0.2in
\end{figure}

However, as shown in \cref{fig:sum_fms}, FM fails to retain PassKey information over a long context. The figure compares M1, 2, and 4 with FM1, 2, and 4 when the memory segment size is 1, 2, and 4 for BSWA and Feedback Memory Segment. TransformerFM outperforms TransformerBSWA, but still falls short of TransformerFAM by a significant margin. In TransformerFM, each activation must possess both local and global representation, similar to TransformerBSWA. However, the absence of an activation specifically responsible for the global representation appears to prevent the retention of critical information like PassKey over an extended context.

Furthermore, TransformerFM exhibited lower performance than TransformerBSWA on GPT-3 tasks (\cref{table:fm_results}). As TransformerFM introduced additional complexity without any discernible performance advantages, we discontinued this architecture. \cref{table:fm_results} presents the results after 25k steps of fine-tuning.

\begin{table}[h]
    
    \centering
    \begin{tabular}{lc ccc ccc}
        \toprule
        \bf Task & \bf Metric & \bf M1 & \bf M2 & \bf M4 & \bf FM1 & \bf FM2 & \bf FM4 \\
        \midrule
        GPT-3 Gen score & EM & 32.5 & \bf 33.6 & 32.9 & 31.9 & 32.0 & 29.7 \\
        GPT-3 Rank & Rank Acc. & 58.5 & \bf 59.0 & 58.5 & 58.2 & 59.0 & 58.3 \\
        Isabelle & Accuracy & 72.5 & 72.2 & 70.9 & 72.6 & \bf 73.5 & 72.3 \\
        PG-19 & Accuracy & 46.4 & 46.1 & 45.9 & 45.8 & \bf 47.2 & 44.5 \\
        XLSum & ROUGE-L & 13.3 & \bf 13.9 & \bf 13.9 & 13.6 & 13.6 & 13.8 \\
        \bottomrule
    \end{tabular}
    \caption{ Comparison of TransformerBSWA and TransformerFM on major tasks }
    \label{table:fm_results}
\end{table}

\subsection{Static Summary Tokens}

As illustrated in \cref{fig:transformer_fam}, "FAM copy" duplicates the previously updated FAM to the current FAM input. However, "FAM copy" is not an essential component in the feedback loop. When FAM compresses the current block, it can also attend to the previous FAM as both key and value, as shown in the experiment in \cref{sssec:num_fam}. Therefore, "FAM copy" is not strictly necessary because past FAM information is propagated through self-attention.

As mentioned in \cref{ssec:init_feedback}, we learn the initial FAM using prompt tuning, and this token can be considered a "Summary" token. An alternative design could employ a persistent summary token for block compression, with the feedback loop enabled by attending to the key-value pairs of previous FAMs.

As illustrated in \cref{fig:sum_fms}, SUM FAM (static summary token) prevents the model from successfully solving the PassKey task. This suggests that the query for summarization needs to be dynamically generated conditioned on the past FAM. A static query may not be able to cover all situations effectively. Additionally, simply attending to the key and value alone does not transmit sufficient information from the past. On the other hand, the "FAM copy" mechanism propagates information from the past FAM to the current FAM through the residual connection of the Transformer, which facilitates better performance on the PassKey task.

\subsection{Diversity Loss}

In \cref{sssec:abl_fam_len}, we observed a performance drop when the FAM length exceeded 64. We hypothesized that FAM was underutilizing its capacity and experiencing mode collapse, where lengthy FAM either focused on a small number of inputs or prevented all inputs from obtaining sufficient information.

As a remedy, we employed diversity loss as an auxiliary loss, similar to Wav2vec 2.0~\cite{baevski2020wav2vec}. Diversity loss aims to maximize the entropy of QK attention probability, as shown in \cref{eq:entropy}. This loss encourages FAM to uniformly attend to all inputs, and all inputs equally attend to FAM. In \cref{eq:div_loss}, $b$ denotes the batch size, $\tau$ represents the block index, $l$ indicates the sequence position within a block, and $h$ refers to multi heads. In \cref{eq:p_bar}, $\bar{p}_{b\tau}$ represents the average attention probability across all sequences and multi-heads for the self-attention of each block. The auxiliary loss functions to regularize this probability towards uniformity.

\begin{subequations}
\begin{align}
\bar{p}_{b\tau} &= \frac{1}{HL} \sum_{h=1}^{H} \sum_{l=1}^{L} p_{b\tau hl} \label{eq:p_bar} \\
\mathcal{L}_{d} &= \frac{1}{B\mathrm{T}} \sum_{b=1}^{B} \sum_{\tau=1}^{\mathrm{T}} - H(\bar{p}_{b\tau}) = \frac{1}{B\mathrm{T}} \sum_{b=1}^{B} \sum_{\tau=1}^{\mathrm{T}} \bar{p}_{b\tau} \log{\bar{p}_{b\tau}} \label{eq:entropy}
\end{align}
\label{eq:div_loss}
\end{subequations}

We trained the model with diversity loss with various weights, but the overall performance was always worse, regardless of the FAM length. It did not help at all even when the FAM length was 256 or longer.

\subsection{Reconstruction Loss}

Compressive transformers~\cite{rae2019compressive} compress memory segments before forwarding them to the next block. The key difference between our approach is that the compressed information is not recurrently connected. The paper proposes a reconstruction loss as an auxiliary loss, which aims to reconstruct the original activations from the compressed activations.

We also experimented with the reconstruction loss. In details, we generated transformer outputs autoregressively from the updated FAM and the query sequence of the original block, and compared them to the original outputs using the MSE loss. However, this did not help to improve the performance.

\section{Experiments details}
\label{app:result}

This section provides additional details about the experiments that were not covered in \cref{sec:exp}.

\subsection{PassKey Retrieval}
\label{app:passkey}

The format of the PassKey retrieval task is shown in \cref{fig:passkey-format}. The original paper~\cite{mohtashami2023landmark} also included a prefix filler, but we removed it in our paper.

\begin{figure}[h]
		{
			
				\texttt{\noindent
					There is an important info hidden inside a lot of irrelevant text. Find it and memorize them. I will quiz you about the important information there.\\
					The pass key is {\color{purple}{<PASS KEY>}}. Remember it. {\color{purple}{<PASS KEY>}} is the pass key.\\
					{\color{purple}{<filler>}}\\
					What is the pass key? The pass key is {\color{purple}{<PASS KEY>}} \\[-2mm]
				}
		}
		\caption{ PassKey Retrieval Format~\cite{mohtashami2023landmark}}
		\label{fig:passkey-format}
\end{figure}

\subsection{Long Context Tasks}
\label{app:lct}

\cref{table:lct} provides detailed information about the long context tasks we used. Due to the limitations of TPU memory, PG-19, Isabelle, and NarrativeQA were truncated to 256k tokens, which is within two standard deviations.

\begin{table}[ht]
    \tiny  % TODO(dongseong)
    % \small
    \centering
    \begin{tabular}{lcccccc}
        \toprule
        \bf Eval task & \bf Metric & \bf Num. & \bf Max len. & \bf Mean len. & \bf \makecell[c]{Standard \\ deviation} & \bf Description \\
        \midrule
        Isabelle & Next Token Accuracy & 16 & 280500 & 60874 & 65008 & Formal theorems \\
        NarrativeQA & ROUGE-L & 5878 & 505949 & 85449 & 84376 & QA on a given underlying narrative \\
        PG-19 & Next Token Accuracy & 50 & 491500 & 94704 & 82677 & Project Gutenberg books published before 1919 \\
        ScrollsQasper & ROUGE-L & 1726 & 24223 & 5027 & 2580 & QA on scientific research papers \\
        ScrollsQuality & Accuracy & 8344 & 9294 & 6220 & 2050 & QA on a given story or document \\
        XLSum & ROUGE-L & 37747 & 13571 & 1888 & 1304 & Multilingual abstractive summarization \\
        \bottomrule
    \end{tabular}
    \caption{ Overview of long-context tasks, including their token count, evaluation metric, and a brief description. The token count is counted after a 256k sentencepiece tokenizer.}
    \label{table:lct}
\end{table}

\cref{table:lct_results} presents the LCT results for TransformerBSWA and TransformerFAM. MX represents the number of memory segments for TransformerBSWA.

\begin{table}[ht]
    
    \centering
    \begin{tabular}{lc cccc ccccc}
        \toprule
        \bf Task & \bf Metric & \bf M0 & \bf M1 & \bf M2 & \bf M3 & \bf M4 & \bf M6 & \bf M7 & \bf M8 & \bf FAM \\
        \midrule
        Isabelle & Accuracy & 65.0 & 72.6 & 72.0 & 70.6 & 70.3 & 69.7 & 67.29 & 68.4 & 73.6 \\
        NarrativeQA & ROUGE-L & 0.0 & 11.1 & 11.8 & 11.2 & 10.0 & 9.7 & 11.0 & 8.9 & 13.5 \\
        PG-19 & Accuracy & 43.0 & 46.4 & 46.6 & 46.3 & 45.8 & 45.7 & 45.39 & 45.7 & 47.0 \\
        ScrollsQasper & ROUGE-L & 2.9 & 5.0 & 4.9 & 4.5 & 4.1 & 4.5 & 5.4 & 5.1 & 7.2 \\
        ScrollsQuality & Accuracy & 27.8 & 26.3 & 26.7 & 26.1 & 26.7 & 26.1 & 25.6 & 26.5 & 27.9 \\
        XLSum & ROUGE-L & 8.5 & 13.6 & 11.8 & 12.4 & 12.3 & 11.8 & 14.2 & 14.4 & 15.9 \\
        \bottomrule
    \end{tabular}
    \caption{ Results of long-context tasks evaluated by Flan 1B with different Transformer models and different memory segment configurations.}
    \label{table:lct_results}
\end{table}

\subsection{GPT-3 Tasks}
\label{app:gpt3}

We evaluated all model sizes on the tasks reported by GPT-3~\cite{brown2020language}. The results are shown in \cref{table:gpt3_raw}.

\begin{table}[ht]
    \small  % TODO(dongseong)
    \centering
    \begin{tabular}{lc cc cc cc}
    \toprule
    \bf Dataset & \bf Metric & \bf BSWA 1B & \bf FAM 1B & \bf BSWA 8B & \bf FAM 8B & \bf BSWA 24B & \bf FAM 24B \\
    \midrule
    GPT-3 Rank & Mean & 60.2 & 61.0 & 72.8 & 74.0 & 78.2 & 78.5 \\
    \midrule
    ANLI R1 & Rank Acc. & 42.2 & 42.5 & 66.1 & 64.8 & 70.8 & 78.3 \\
    ANLI R2 & Rank Acc. & 37.7 & 37.8 & 50.6 & 50.3 & 59.6 & 65.2 \\
    ANLI R3 & Rank Acc. & 39.4 & 38.6 & 50.5 & 51.3 & 60.1 & 62.8 \\
    ARC Challenge & Rank Acc. & 37.6 & 37.9 & 57.0 & 56.6 & 62.8 & 63.9 \\
    ARC Easy & Rank Acc. & 69.1 & 69.7 & 81.6 & 81.6 & 87.6 & 87.2 \\
    BoolQ & Rank Acc. & 74.7 & 74.2 & 85.6 & 85.8 & 89.1 & 89.9 \\
    CB & Rank Acc. & 73.2 & 78.6 & 67.9 & 91.1 & 100.0 & 82.1 \\
    COPA & Rank Acc. & 71.0 & 78.0 & 90.0 & 91.0 & 91.0 & 93.0 \\
    HellaSwag & Rank Acc. & 55.6 & 56.8 & 76.1 & 76.5 & 83.0 & 83.1 \\
    MultiRC & Rank Acc. & 70.2 & 68.0 & 80.2 & 80.8 & 85.2 & 85.1 \\
    OpenbookQA & Rank Acc. & 49.8 & 51.6 & 57.6 & 58.8 & 64.4 & 66.6 \\
    PIQA & Rank Acc. & 74.1 & 74.5 & 80.4 & 81.0 & 84.2 & 84.0 \\
    RACE-H & Rank Acc. & 39.3 & 38.8 & 49.7 & 50.5 & 55.1 & 56.3 \\
    RACE-M & Rank Acc. & 55.0 & 53.6 & 65.8 & 65.9 & 70.8 & 70.5 \\
    ReCoRD & Rank Acc. & 80.3 & 80.7 & 89.5 & 89.6 & 91.2 & 90.0 \\
    RTE & Rank Acc. & 63.2 & 65.0 & 87.0 & 87.0 & 86.3 & 87.7 \\
    StoryCloze & Rank Acc. & 73.8 & 75.2 & 82.7 & 82.6 & 87.5 & 85.8 \\
    WiC & Rank Acc. & 53.4 & 51.9 & 58.6 & 62.2 & 54.5 & 53.3 \\
    Winograd & Rank Acc. & 73.3 & 74.4 & 88.6 & 85.7 & 88.3 & 89.0 \\
    Winogrande & Rank Acc. & 61.1 & 61.9 & 78.2 & 77.5 & 82.3 & 84.1 \\
    WSC273 & Rank Acc. & 70.9 & 72.3 & 85.3 & 84.2 & 87.4 & 89.5 \\
    \midrule
    GPT-3 Gen & Mean & 33.9 & 34.7 & 54.3 & 54.9 & 62.6 & 63.4 \\
    \midrule
    LAMBADA & Decode Acc. & 63.0 & 65.1 & 81.0 & 81.2 & 83.4 & 83.9 \\
    Natural Questions & EM score & 10.8 & 10.7 & 25.6 & 25.3 & 37.6 & 37.3 \\
    SQuADv2.0 & EM score & 44.7 & 45.7 & 68.9 & 72.2 & 79.1 & 79.2 \\
    TriviaQA & EM score & 24.3 & 24.6 & 58.3 & 58.2 & 69.7 & 76.1 \\
    WebQuestions & EM score & 26.5 & 27.4 & 37.8 & 37.5 & 43.3 & 40.5 \\
    \bottomrule
    \end{tabular}
\caption{
Comparison of performance of GPT-3 tasks~\cite{brown2020language} between BSWA and FAM across 1B, 8B, and 24B models.
}
\label{table:gpt3_raw}
\end{table}

\subsection{Complexity}
\label{ssec:comp}

BSWA and FAM have memory and computational complexity of $O(L \times C)$, where $C$ is the chunk size. \cref{table:mem} and \cref{table:speed} show memory and inference time through inference jobs on TPUv4. The most tokens in the experiment are pre-filled and generation is limited to 256 tokens.

\begin{table}[ht]
    \centering
    \begin{tabular}{lccccc}
        \toprule
        \bf Memory (GB) & \bf 26k & \bf 34k & \bf 66k & \bf 130k & \bf 258k \\
        \midrule
        BSWA & 4.4 &4.8 & 6.4 & 9.6 & 16.0 \\
        FAM & 4.5 & 4.9 & 6.5 & 9.7 & 16.1 \\
        \bottomrule
    \end{tabular}
    \caption{Memory Usage Comparison: BSWA vs. FAM across varying sequence lengths.}
    \label{table:mem}
\end{table}

\begin{table}[ht]
    \centering
    \begin{tabular}{lccccc}
        \toprule
        \bf Process secs & \bf 26k & \bf 34k & \bf 66k & \bf 130k & \bf 258k \\
        \midrule
        BSWA & 16.2 & 20.9 & 40.2 & 76.8 & 154.1 \\
        FAM & 16.5 & 21.2 & 40.7 & 77.2 & 154.9 \\
        \bottomrule
    \end{tabular}
    \caption{Processing seconds Comparison: BSWA vs. FAM across varying sequence lengths.}
    \label{table:speed}
\end{table}

\section{Related Work}
\label{app:related}

The Transformer architecture exhibits a quadratic complexity with respect to context length, a significant limitation. To address this, several research works have focused on approximating the attention mechanism.  One approach involves sparse attention, where only a subset of important tokens are attended to, as seen in models like Sparse Transformer~\cite{child2019generating}, Big Bird~\cite{zaheer2020big}, Reformer~\cite{kitaev2020reformer}, Routing Transformer~\cite{roy2021efficient}, and TOVA~\cite{oren2024transformers}.  Linear approximation methods offer an alternative, seeking to circumvent the quadratic complexity by altering attention calculations, as exemplified by Linformer~\cite{wang2020linformer}, Linear Transformer~\cite{katharopoulos2020transformers}, Performer~\cite{choromanski2020rethinking}, and Nyströmformer~\cite{xiong2021nystromformer}.  Finally, some research explores entirely different sequence-to-sequence architectures as replacements for attention-based Transformers, including MLP-mixer~\cite{tolstikhin2021mlp}, State Space Models~\cite{rangapuram2018deep}, S4~\cite{gu2021efficiently}, Mamba~\cite{gu2023mamba}, and RWKV~\cite{peng2023rwkv}.

Global Workspace Theory (GWT)~\cite{baars2005global} is a leading theory of consciousness. According to GWT, the human brain possesses a global workspace where various modules, such as sensory input, memories, and internal representations, converge. The attention mechanism in brain acts as a spotlight, focusing on specific inputs among the multitude, and transforming this unconscious activity into conscious awareness. These "momentarily active, subjectively experienced" events are then stored in working memory. TransformerFAM draws inspiration from GWT, adopting its principle of a unified attention mechanism for processing homogenous, heterogeneous, and feedback data.

\section{Attention Visualization}
\label{app:visualization}

\begin{figure*}[t]
    \centering
    \begin{subfigure}{.33\textwidth}
    \includegraphics[width=\linewidth]{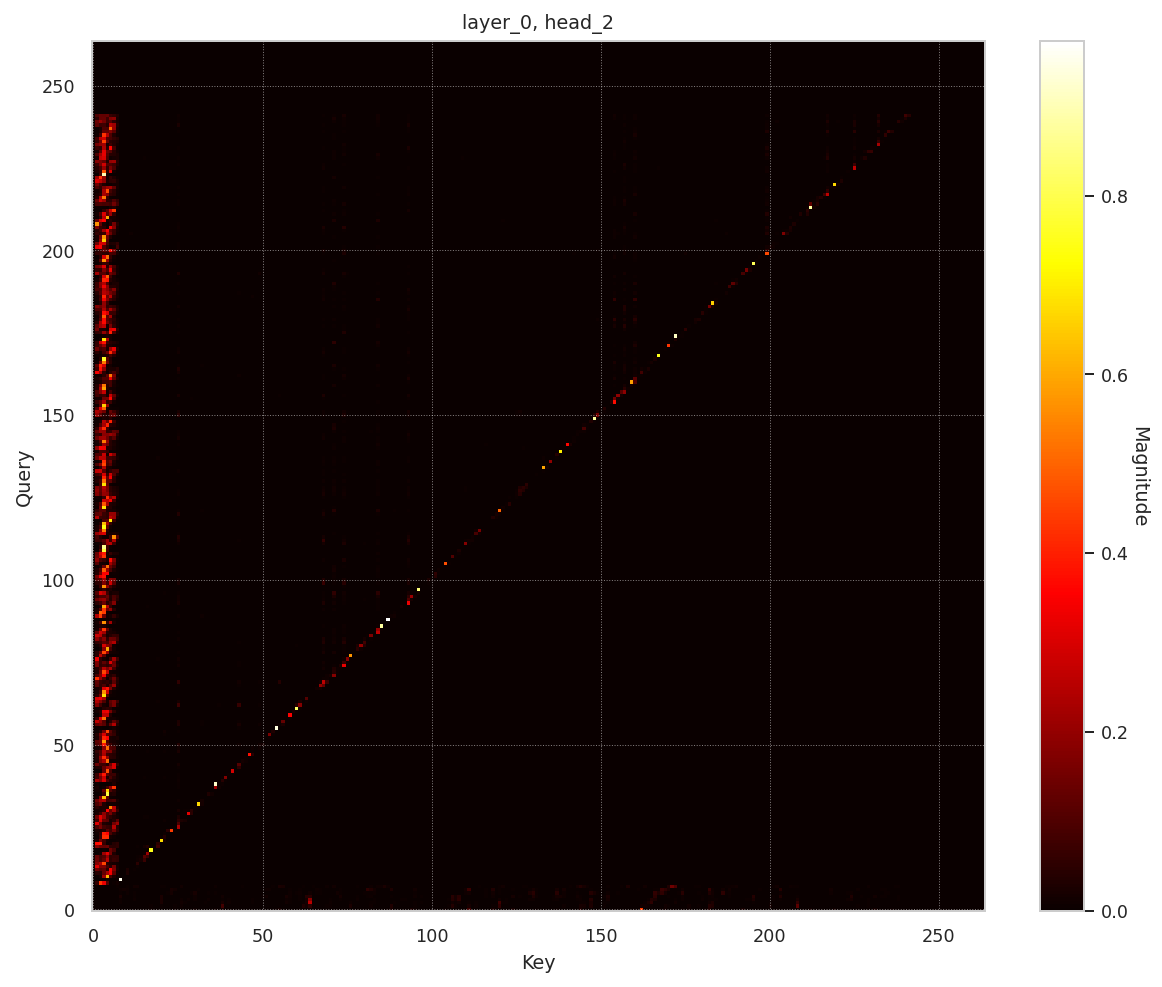}
    \caption{0th Layer, 2nd Head}
    \end{subfigure}\hfill
    \begin{subfigure}{.33\textwidth}
    \includegraphics[width=\linewidth]{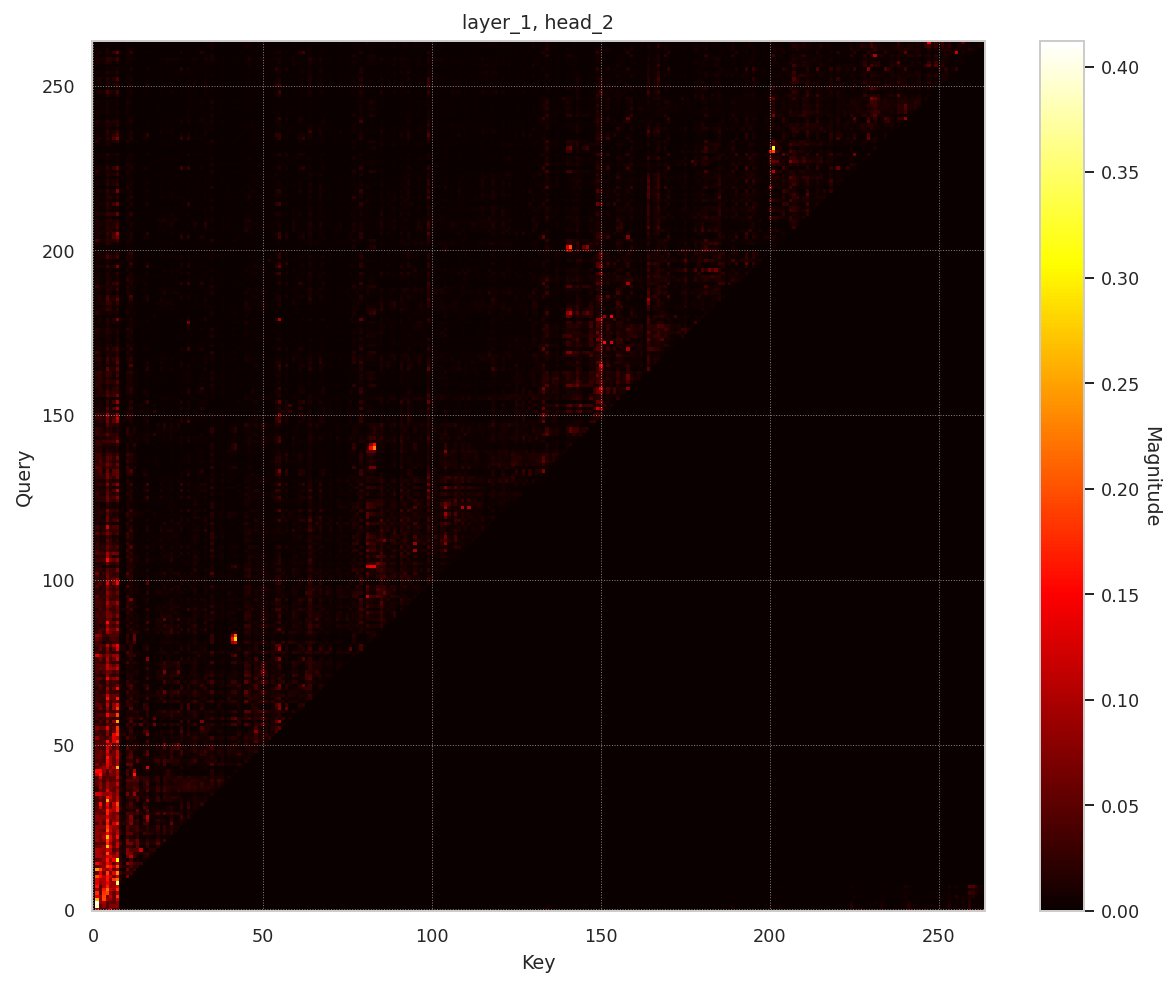}
    \caption{1th Layer, 2nd Head}
    \end{subfigure}\hfill
    \begin{subfigure}{.33\textwidth}
    \includegraphics[width=\linewidth]{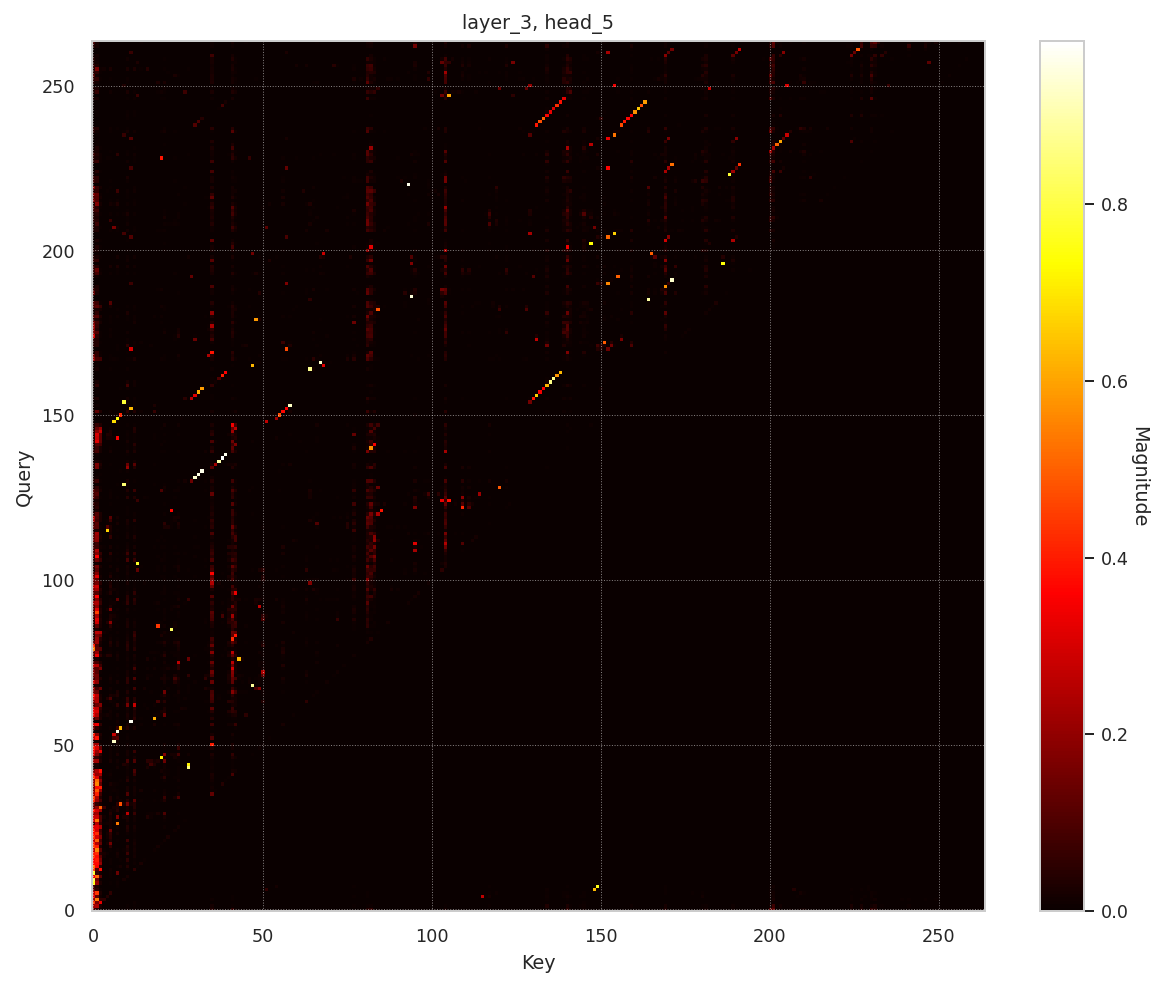}
    \caption{3rd Layer, 5th Head}
    \end{subfigure}\hfill
    \begin{subfigure}{.33\textwidth}
    \includegraphics[width=\linewidth]{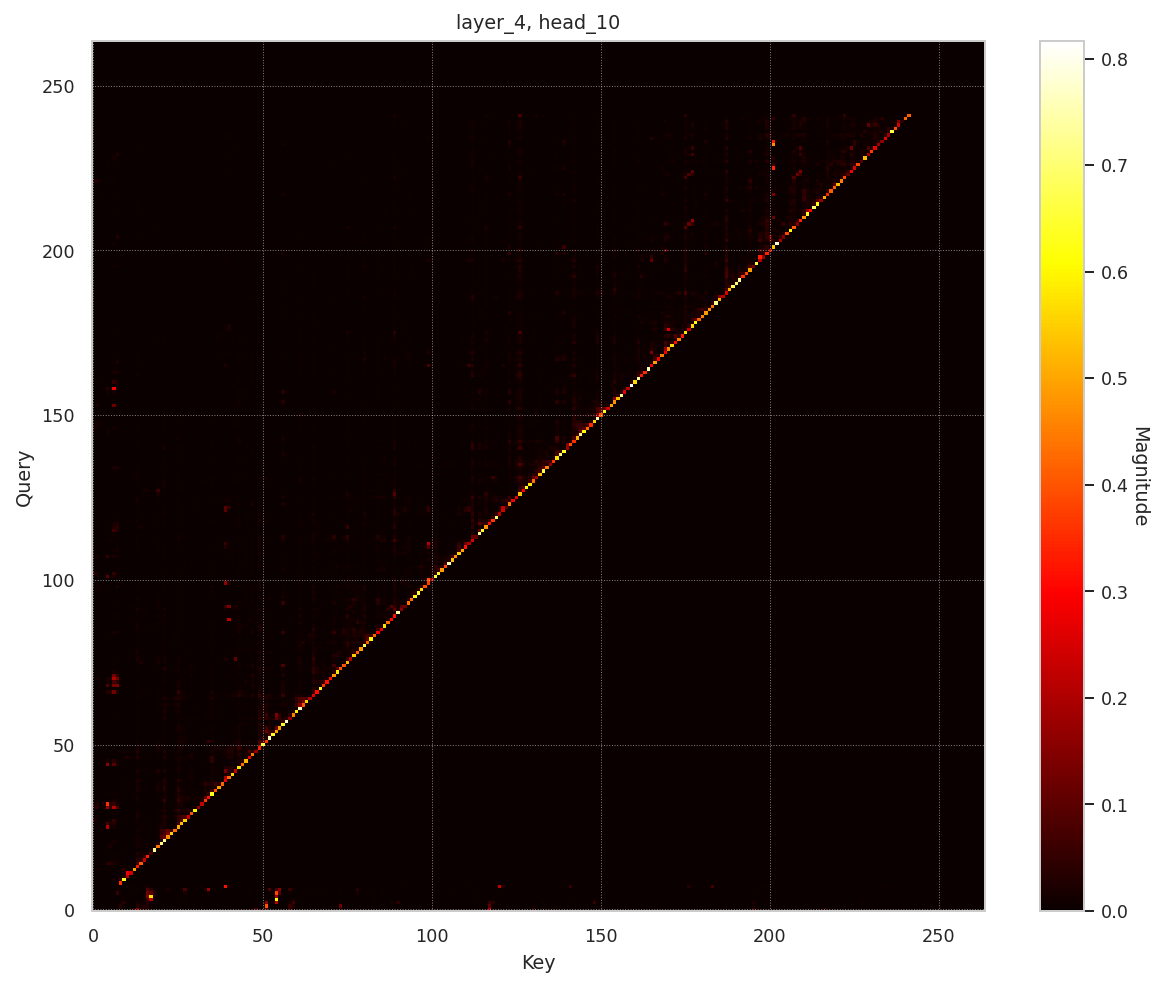}
    \caption{4th Layer, 10th Head}
    \end{subfigure}\hfill
    \begin{subfigure}{.33\textwidth}
    \includegraphics[width=\linewidth]{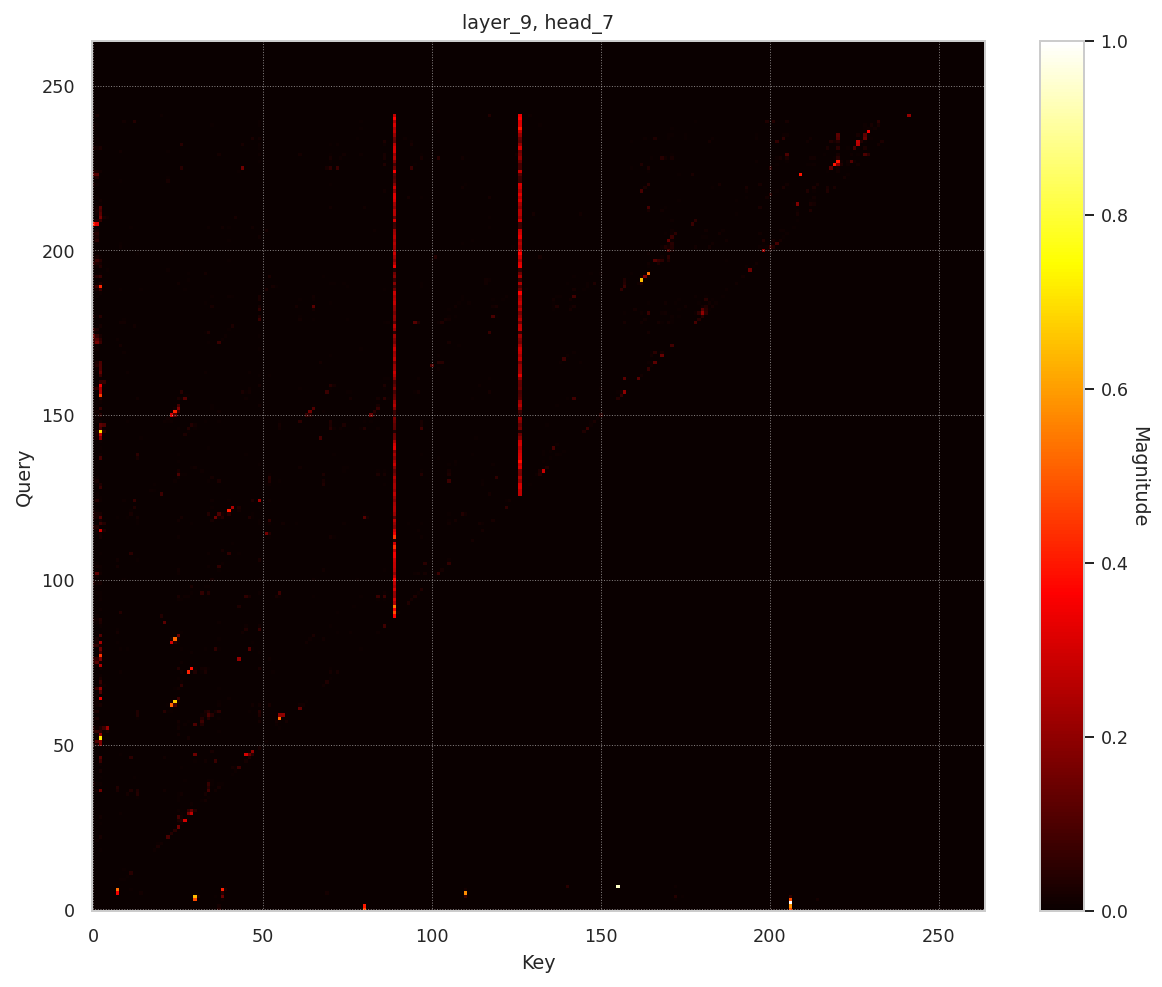}
    \caption{9th Layer, 7th Head}
    \end{subfigure}\hfill
    \begin{subfigure}{.33\textwidth}
    \includegraphics[width=\linewidth]{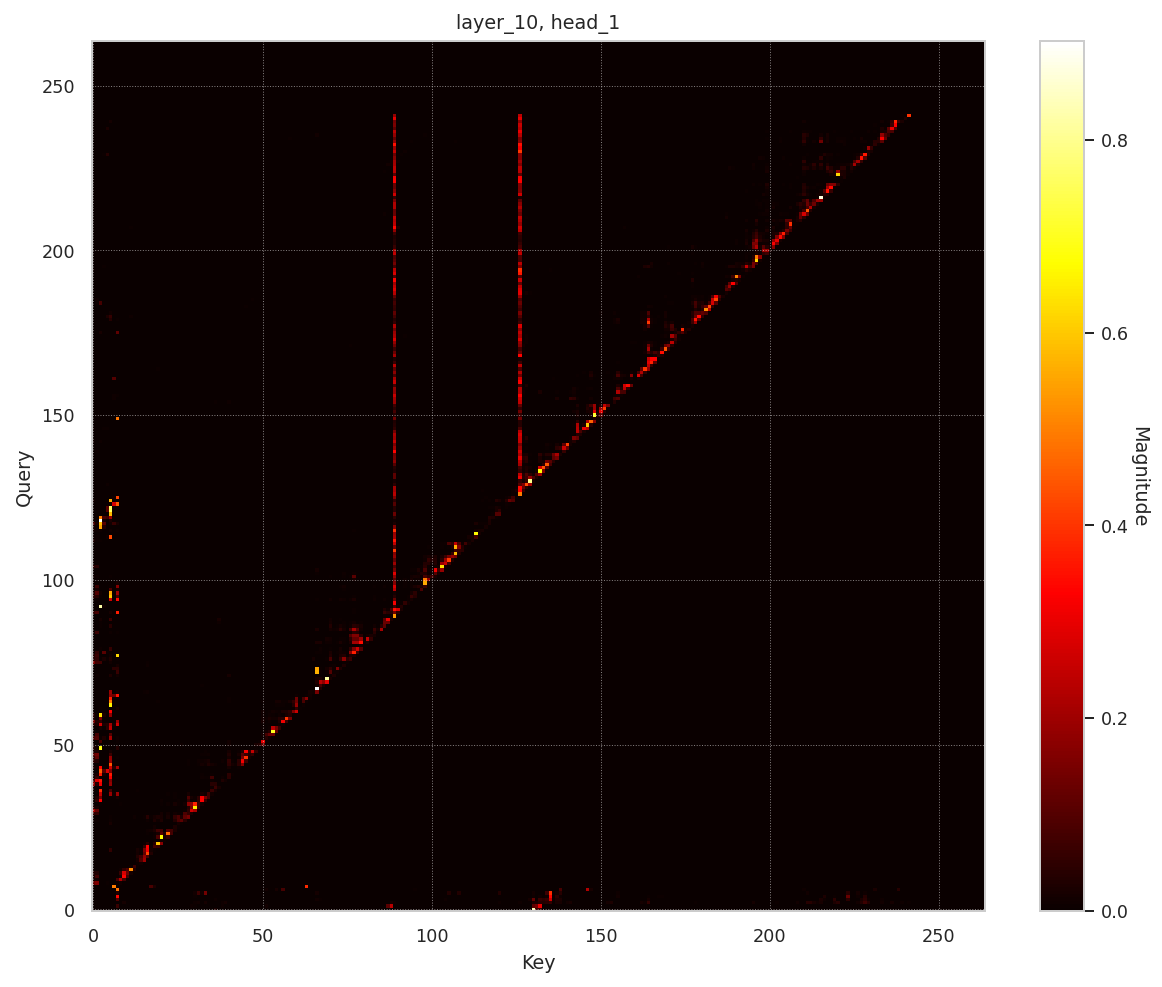}
    \caption{10th Layer, 1st Head}
    \end{subfigure}\hfill
    \begin{subfigure}{.33\textwidth}
    \includegraphics[width=\linewidth]{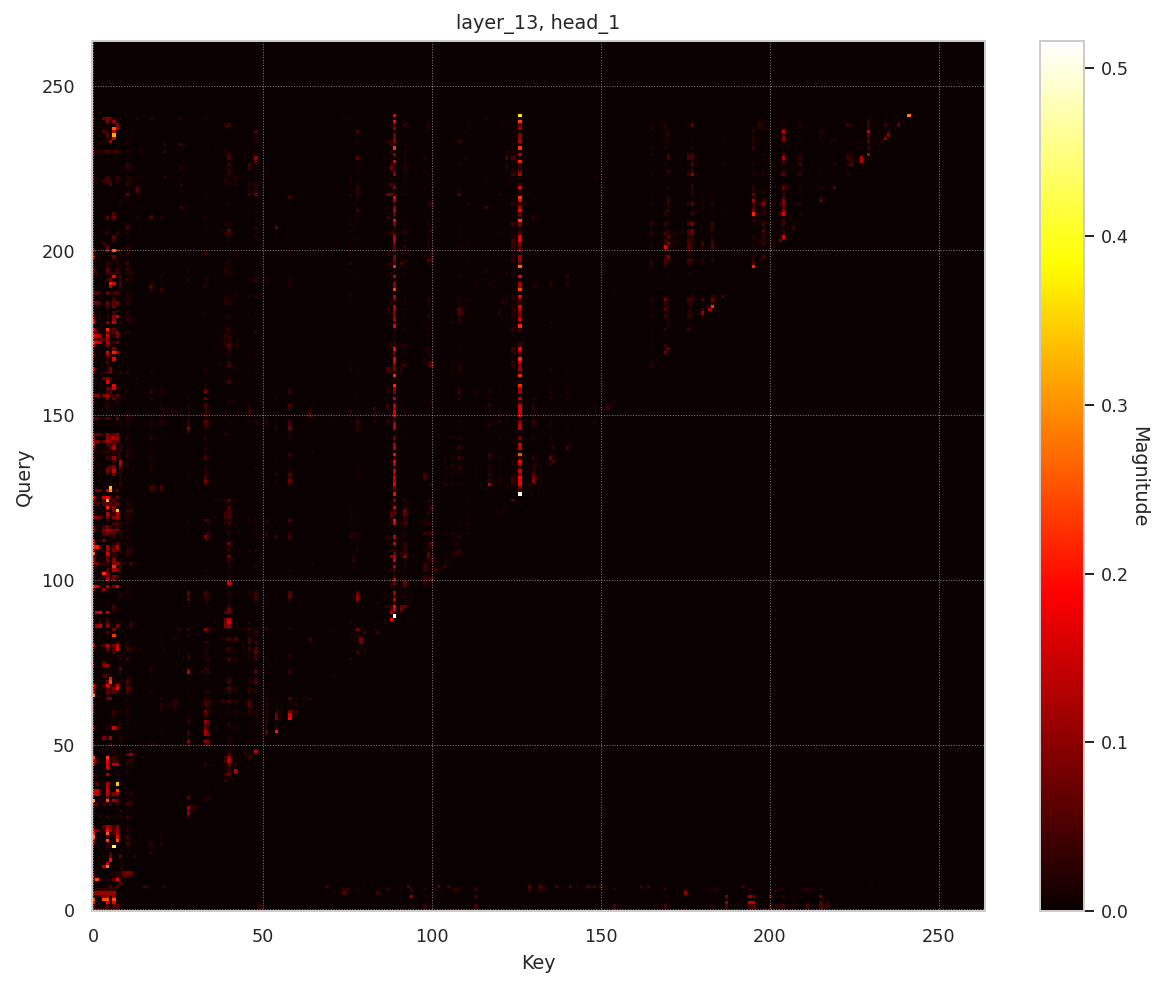}
    \caption{13th Layer, 1st Head}
    \end{subfigure}\hfill
    \begin{subfigure}{.33\textwidth}
    \includegraphics[width=\linewidth]{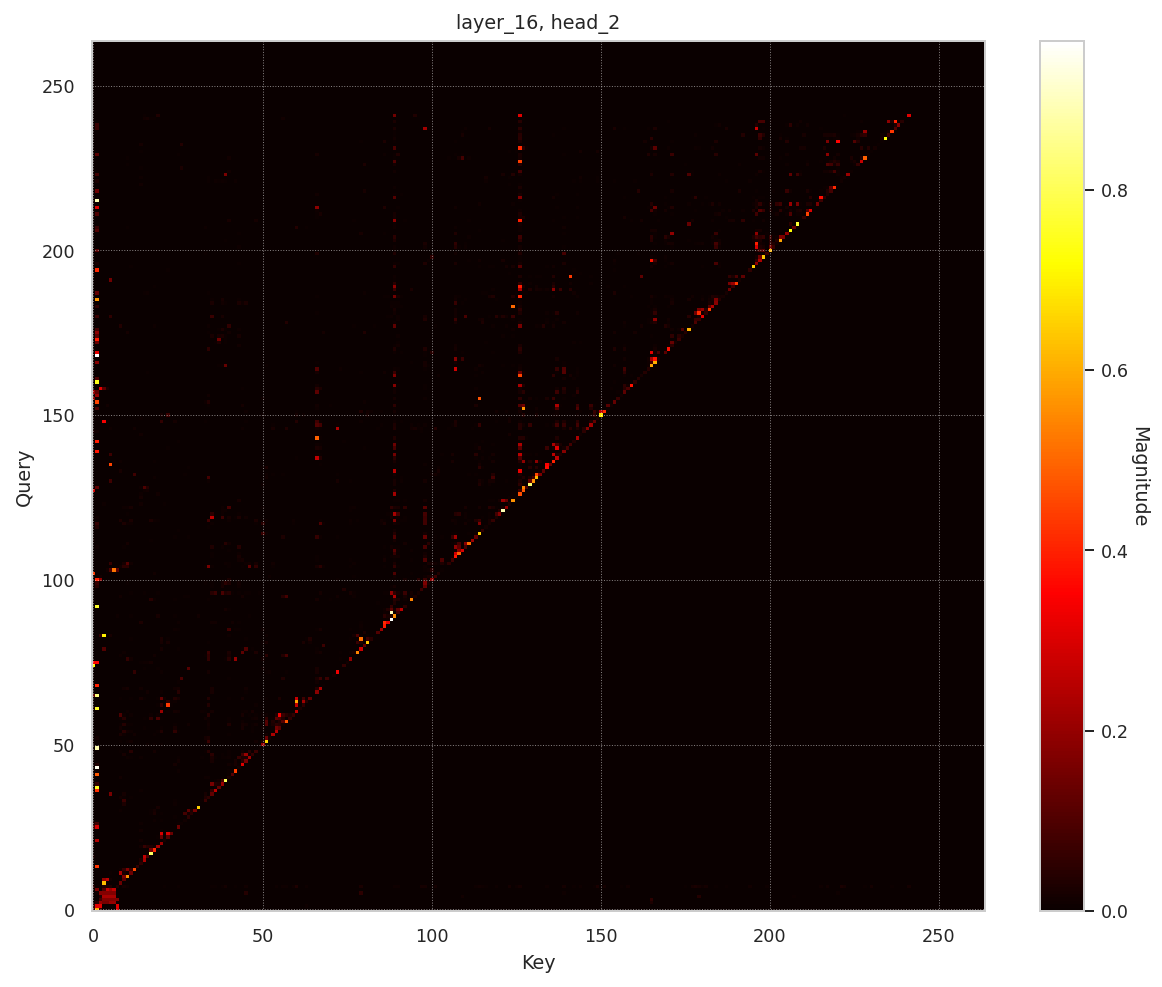}
    \caption{16th Layer, 2nd Head}
    \end{subfigure}\hfill
    \begin{subfigure}{.33\textwidth}
    \includegraphics[width=\linewidth]{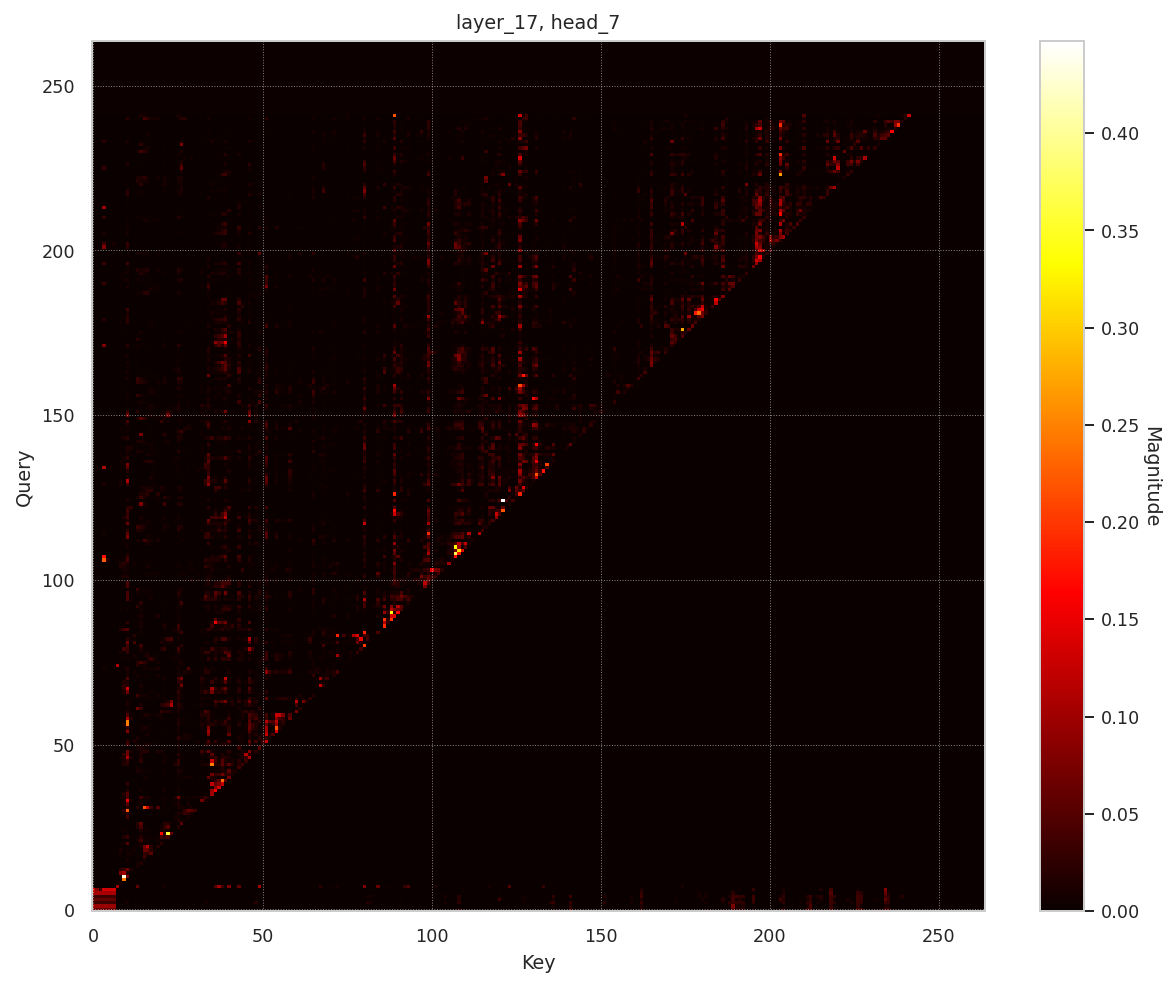}
    \caption{17th Layer, 7th Head}
    \end{subfigure}\hfill
    \caption{ Selected attention map of TransformerFAM with block size of 256 and FAM length of 8. The vertical axis of the attention map represents the Query, and the horizontal axis represents the Key. The FAM is prepended, and is located at the bottom left corner.}
    \label{fig:visual}
    % \vspace{-4mm}
\end{figure*}

\cref{fig:visual} depicts the attention map for each head in each layer of a 1B model. FAM is prepended and is located at the bottom left corner. The bright spots along the left edge represent the block inputs attending to FAM, while the bright spots along the bottom edge represent FAM compressing the corresponding block. Overall, the block inputs actively reference FAM, while FAM compresses only the selective inputs.

% go/xformer-fam-visualization-colab

\section{Limitations}
\label{app:limit}

While the results presented in Table \ref{table:8b_lct} demonstrate that TransformerFAM shows improvements on long-context tasks, these gains are not yet substantial, highlighting the need for further development and refinement of working memory mechanisms.

In this paper, we have taken an initial step towards integrating attention-based working memory, a concept inspired by neuroscience, into deep learning architectures. We believe there is significant potential for further exploration in this direction, and we encourage future research to continue addressing the ongoing challenge of limited memory in deep learning models.

\section{Broader Impacts}
\label{app:impact}

While our work is inspired by the concept of working memory from neuroscience, as discussed in Section \cref{ssec:lct}, achieving a human-level implementation remains a significant challenge. This research represents an initial step in that direction.

The potential societal impacts of such advanced working memory in LLMs could be substantial, with applications like highly personalized AI assistants. However, these impacts are currently speculative due to the nascent stage of this research.

In the immediate future, the primary benefit of our work is expected to be improvements in the efficiency and effectiveness of LLMs, with potential applications across various domains such as education, healthcare, and communication. As with any technology, we recognize the possibility of misuse and encourage ongoing research into the ethical implications of LLMs and related advancements in artificial intelligence.